\documentclass[sigconf]{acmart}

\usepackage{booktabs}
\usepackage{verbatim} 
\usepackage{mathtools} 
\usepackage{appendix}

\usepackage{amsmath,amsfonts}
\usepackage{pifont}
\usepackage{algorithm}
\usepackage{algpseudocode}
\usepackage{float}
\usepackage{array}
\usepackage{graphicx}
\usepackage{subcaption}
\usepackage{cleveref}
\usepackage{enumitem}
\usepackage{multirow}
\usepackage{amsthm}
\usepackage{xspace}

\newtheorem{definition}{Definition}

\newcommand{\datasetFont}{\text}
\newcommand{\ours}{\datasetFont{EdgeRec3D}\xspace}
\newcommand{\cmark}{\ding{51}}%
\newcommand{\xmark}{\ding{55}}%

\title{Personalized Product Assortment with Real-time 3D Perception and Bayesian Payoff Estimation}

\author{Porter Jenkins*, Michael Selander$^{\dagger}$, J. Stockton Jenkins$^{\dagger}$, Andrew Merrill$^{\dagger}$, Kyle Armstrong$^{\dagger}$}
\affiliation{\institution{Brigham Young University*, Delicious AI$^{\dagger}$}\country{United States}}
\email{pjenkins@cs.byu.edu*}
\email{{michael.selander, stockton.jenkins, andrew.merrill, kyle.armstrong}@deliciousai.com}

\copyrightyear{2024}
\acmYear{2024}
\setcopyright{acmlicensed}\acmConference[KDD '24]{Proceedings of the 30th ACM SIGKDD Conference on Knowledge Discovery and Data Mining}{August 25--29, 2024}{Barcelona, Spain}
\acmBooktitle{Proceedings of the 30th ACM SIGKDD Conference on Knowledge Discovery and Data Mining (KDD '24), August 25--29, 2024, Barcelona, Spain}
\acmDOI{10.1145/3637528.3671518}
\acmISBN{979-8-4007-0490-1/24/08}

\begin{document}

\begin{abstract}
    Product assortment selection is a critical challenge facing physical retailers. Effectively aligning inventory with the preferences of shoppers can increase sales and decrease out-of-stocks. However, in real-world settings the problem is challenging due to the combinatorial explosion of product assortment possibilities. Consumer preferences are typically heterogeneous across space and time, making inventory-preference alignment challenging. Additionally, existing strategies rely on syndicated data, which tends to be aggregated, low resolution, and suffer from high latency. To solve these challenges, we introduce a real-time recommendation system, which we call \ours. Our system utilizes recent advances in 3D computer vision for perception and automatic, fine grained sales estimation. These perceptual components run on the edge of the network and facilitate real-time reward signals. Additionally, we develop a Bayesian payoff model to account for noisy estimates from 3D LIDAR data. We rely on spatial clustering to allow the system to adapt to heterogeneous consumer preferences, and a graph-based candidate generation algorithm to address the combinatorial search problem. We test our system in real-world stores across two, 6-8 week A/B tests with beverage products and demonstrate a 35\% and 27\% increase in sales respectively. Finally, we monitor the deployed system for a period of 28 weeks with an observational study and show a 9.4\% increase in sales. 
\end{abstract}
\maketitle

\section{Introduction}

The problem of effectively allocating shelf space is one of the most critical decisions facing physical retailers \cite{yang1999study}. Optimal placement of physical product increases sales and decreases excess inventory. Prior work has shown that good product allocation can call the consumer's attention and encourage ``impulse'' purchases \cite{hwang2005model, Mattila, badgaiyan2015does}. In most real-world scenarios, the number of available products far exceeds the amount of available shelf space. Additionally, the retailer must consider varying attributes of products such as price,  shape, and dynamic consumer preferences. As such, the product assortment problem is a combinatorial decision problem. 


A variety of previous work seeks to solve the shelf space allocation problem. Classical optimization methods such as integer \cite{yang1999study, geismar2015maximizing} linear, and dynamic \cite{zufryden} programming, genetic algorithms \cite{ZHENG2023101251, castelli2014genetic, urban1998inventory} and simulated annealing approaches \cite{borin1994model} have been extensively studied. More recently, the Data Mining field has sought to replace the formal space elasticity optimization problem with association rule mining \cite{tatiana2018market, pandit2010intelligent, chen2007data, brijs1999using}, and collaborative filtering \cite{park2019group} as they can better leverage large-scale databases to generate effective product sets. In practice, these methods typically rely on syndicated data \footnote{Syndicated data refers to retail sales data that is aggregated across many stores and sold by third parties} as input. These datasets tend to be aggregated, of low resolution, and can take weeks or months to assemble. In contrast, this work offers a solution to produce similar data instantly, in the hand of the person who needs it.  A real-time approach to the problem is critical because it closes the feedback loop and facilitates faster discovery and convergence to shopper preferences.





\begin{figure*}
     \centering
     \begin{subfigure}[b]{0.4\textwidth}
         \centering
         \includegraphics[width=\textwidth]{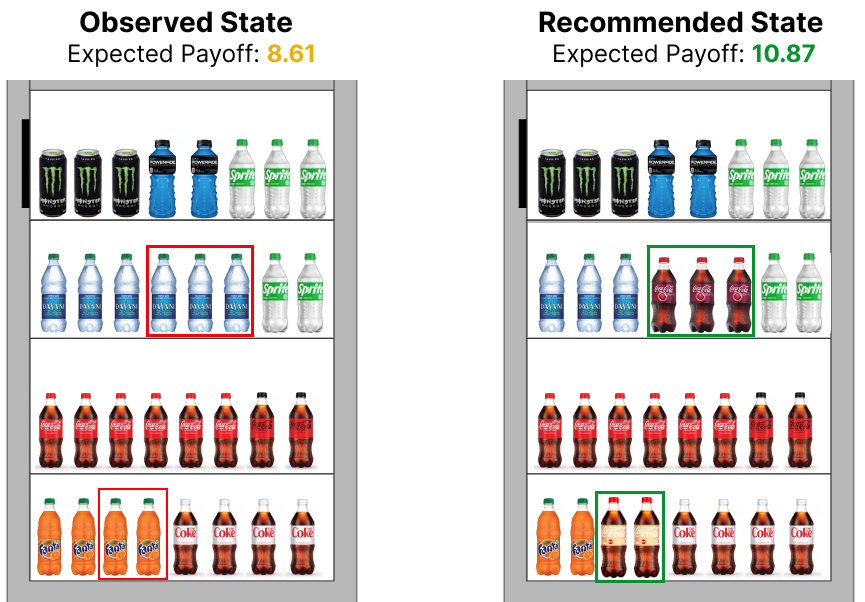}
         \caption{Display states: before and after recommendation}
         \label{fig:intro_a}
     \end{subfigure}
     \hfill
     \begin{subfigure}[b]{0.25\textwidth}
         \centering
         \includegraphics[width=\textwidth]{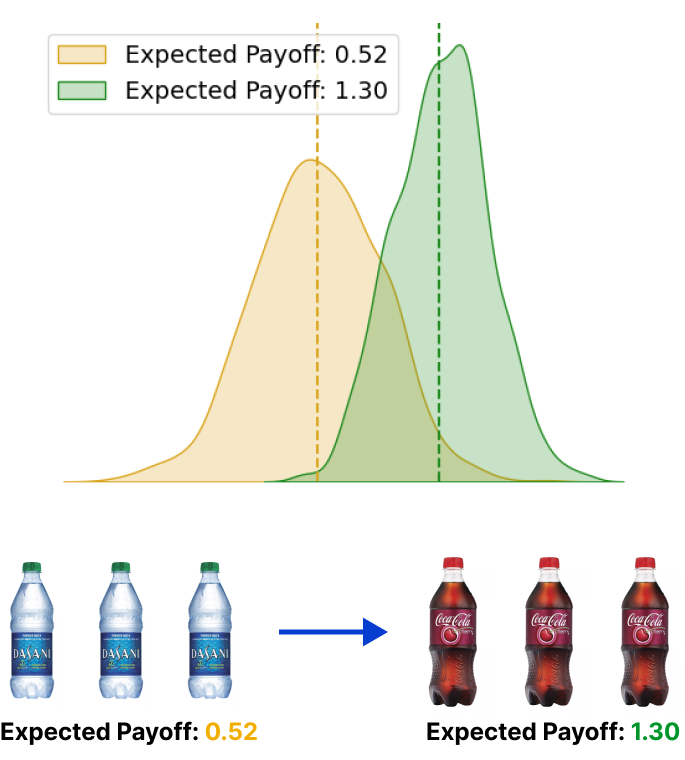}
         \caption{Recommended Change \#1}
         \label{fig:intro_b}
     \end{subfigure}
     \hfill
     \begin{subfigure}[b]{0.25\textwidth}
         \centering
         \includegraphics[width=\textwidth]{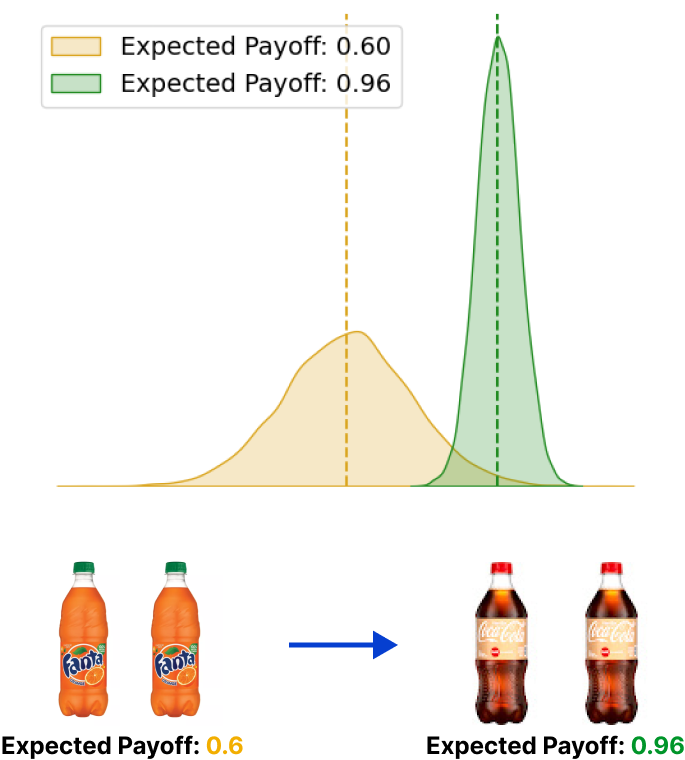}
         \caption{Recommended Change \#2}
         \label{fig:intro_c}
     \end{subfigure}
     \caption{The product assortment problem can be described as choosing the set of products, and corresponding quantities, that will maximize expected payoff, subject to the total number of discrete locations (slots) within the product display. Our system  maximizes expected payoff, while accounting for uncertainty. } 
     \label{fig:into}
     \vspace{-5mm}
\end{figure*}


The problem of real-time product assortment recommendation has four core challenges.  First, obtaining sales estimates in real-time is not possible under current methodologies. Second, the problem is a special case of the knapsack problem \cite{dudzinski1987exact, yang1999study}, a well-known combinatorial optimization problem, and is therefore  NP-complete. The number of possible combinations grows factorially, and follows the equation:

\begin{equation}\label{eq:factorial}
    c =\binom{p + m -1}{m} = \frac{(p+m -1)!}{m!(p-1)!}
\end{equation}

for $p$ products and $m$ discrete product locations, or slots. Even in somewhat small sub-problems comprised of a shelf with 20 slots and 100 products the number of possible combinations is $2.5 \times 10^{22}$. Second, most retail outlets are dynamical systems. The set of products on display frequently changes to account for out-of-stocks or supply chain constraints. Third, preferences are heterogeneous across both time and space \cite{allenby1998marketing, rossi2003bayesian, andersen2010preference, jedidi1997finite}. Limited time products are periodically introduced into the market to satisfy seasonal demand. Additionally, product space allocation should be targeted to meet preferences that vary with respect to demographics. For example, a store near a college campus might sell more energy drinks than a store with an older population of customers. Classical methods are unable to account for these issues.




To solve these challenges, we design a large-scale recommender system, which we call \ours. The system guides retail workers and recommends the appropriate product set to put on display. A key differentiator of our system is that we use on-edge, 3D computer vision models \cite{jenkins2023wacv} for real-time perception and automatic sales estimation. This real-time perceptual layer is implemented at the edge of the network, making it easy to implement the recommended set quickly, and closes the feedback loop under changing market dynamics. A major trade-off under this design is the introduction of noise into estimates of sales. To quantify our uncertainty of product sales due to this observation noise, we propose a Robust Bayesian Payoff (RBP) reward model that learns the posterior distribution of product payoff \footnote{In this paper, we use the terms \textit{payoff} and \textit{reward} interchangeably}. We address the combinatorial search problem with intelligent candidate generation, along with a simple heuristic search based on an uncertainty penalized ranking statistic, which discounts uncertain recommendations. A novel spatial clustering technique along with the hierarchical parameterization of RBP allow for targeted recommendations and help solve the preference heterogeneity problem. 

We validate our recommender system by running both offline and online experiments. Offline, \ours outperforms classic methods such as Linear Programming, Dynamic Programming, $\epsilon$-greedy, Genetic algorithms and modern techniques such as deep ensembles and model-based offline RL. Moreover, we perform two real-world A/B tests, each lasting 6-8 weeks. In these two controlled experiments, we demonstrate a 35\% and 27\% increase in product-level sales, respectively. Finally, a 28 week deployment study results in a 9.4\% increase in product-level sales. Our key contributions can be summarized by the following:
\begin{itemize}
    \item We propose \ours, a real-time, 3D computer vision recommender system to solve the shelf space allocation problem. The system implements 3D visual perception models embedded in mobile devices to observe product display states and reward (sales) in real-time.
    \item \ours uses a novel Robust Bayesian Payoff (RBP) model to estimate action payoffs, and quantify uncertainty under observation noise. This uncertainty quantification informs a heuristic search algorithm to help solve combinatorial explosion. Additionally, we devise a novel spatial clustering method, called SpAGMM, to produce store-level demographic profiles and solve the preference heterogeneity problem.
\end{itemize}

We believe this work will broadly be interesting to those working in real-world recommender systems, brick and mortar commerce, and data-driven marketing.
\section{Preliminaries}\label{sec:prelim}


 The product assortment problem is a special case of the knapsack problem \cite{yang1999study}.

\begin{definition}[Product Assortment Problem] We consider the problem of choosing from a set of $i=1, ..., N$ products with quantities, $q_i \in \{0, 1, 2, ...\}$. The quantity, $q_i$, denotes the number of discrete product locations (slots) occupied by product $i$, on a product display $d_j$. We let $p_i \in \{0, 1\}$ be an indicator variable denoting the selection status of product $i$ (also called the 0-1 knapsack \cite{balas1980algorithm}). Each store, $s_l$, is expected to contain multiple displays, $s_l = \{d^{(l)}_1, d^{(l)}_2, ... d^{(l)}_j\}$. Additionally, each display has a maximum capacity, $M^{(l)}_j$, denoting the available product volume. We assume reward follows an unknown, store-conditional distribution function, $r_{ij} \sim p(r_{ij} | s_l, q_{ij})$ that varies by quantity allocated. For example, one might expect diminishing marginal returns for larger quantities of the same product. Therefore our goal is to maximize expected reward for each display, $d_j$, at each discrete time step, $t$

{\small
\begin{flalign}
     \textrm{maximize} \sum_{i}^{N} \mathbb{E}[r_{ij}|s_l, q_{ij}] p_{ij}  \\
    \textrm{subject to} \sum_{i}^{N} p_{ij} q_{ij} \leq M_j \
\end{flalign}
}

\end{definition}

One key challenge in the Product Assortment Problem is learning the store-conditional reward function, $r_i \sim p(r_{ij} | s_l, q_{ij})$, since preferences tend to be heterogeneous across locations. We tackle this problem by learning demographic profiles for stores and clustering them to estimate   $\mathbb{E}[r_{ij} | s_l, q_{ij}]$.
\begin{definition}[Demographic Spatial Clustering Problem] 

Given a set of spatial areas, $\alpha_z \in \{\alpha_z: 1 \leq z \leq Z \} $, each with a demographic vector, $x_z \in \mathbb{R}^b$ and stores, $s_l \in  \{s_l: 1 \leq l \leq L\}$, we have two goals. First, match stores to a set of areas by constructing a neighborhood around each store, $s_l$: $\mathcal{N}{s_l}(\{(a_1, x_1), (a_2, x_2), ...\})$. Second within each neighborhood, produce a store-level demographic profile by aggregating over the demographic vectors, $X_l = g\big[\mathcal{N}{s_l}(\{(a_1, x_1),  ...\})\big]$. The aggregation function, $g(\cdot)$ could be any scalar-valued aggregation (i.e., mean, median, weighted average). Finally, we cluster the set of $L$ store demographic profiles, $X_l$'s, into $K$ different clusters. We expect stores within each cluster to have similar preferences, making it easier to estimate the store-conditional expectation, $\mathbb{E}[r_{ij} | s_l, q_{ij}]$.
\end{definition}

A summary of notation used throughout the paper is presented Appendix \ref{sec:notation}.

\section{Method}\label{sec:method}

In the following section we discuss the various components of \ours. Our recommendation engine is comprised of the following modular components: 3D data collection and sales inference, spatial clustering, candidate generation, value estimation, ranking and heuristic search. The output is a recommended product set tailored \textit{to each display location} in a store, $d_{ij}$, which a retail worker can choose to physically implement.

\begin{figure*}
     \centering
     \begin{subfigure}[b]{0.75\textwidth}
         \centering
         \includegraphics[width=\textwidth]{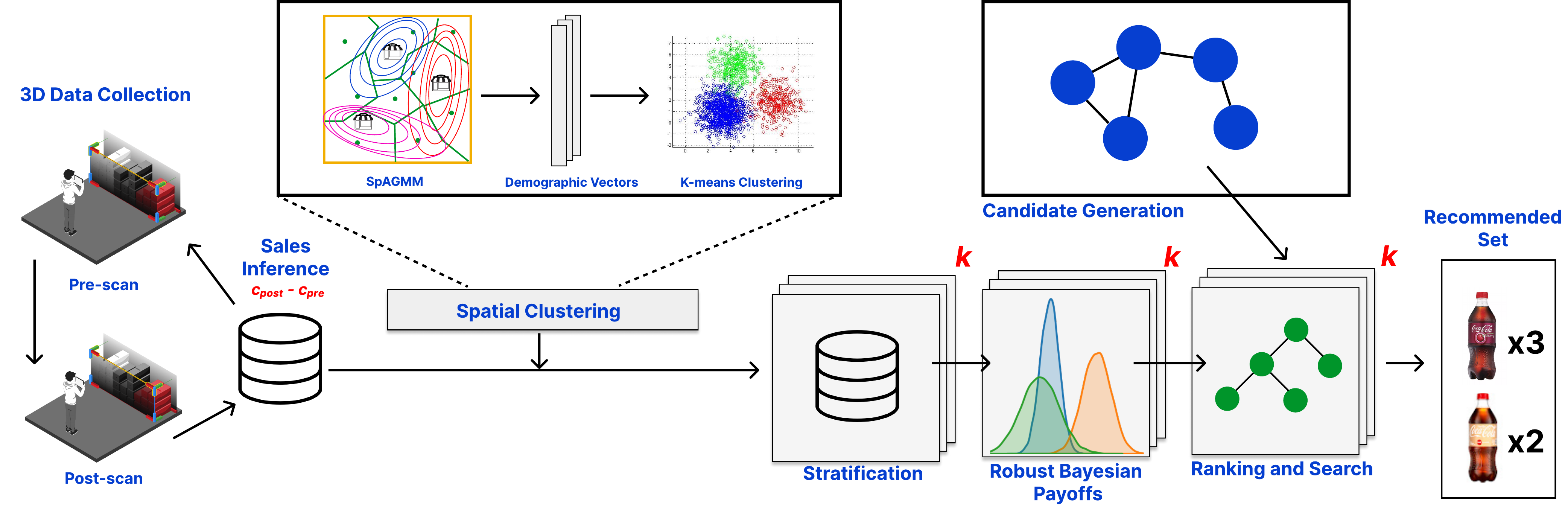}
         \caption{Logical architecture of \ours}
         \label{fig:arch_a}
     \end{subfigure}
     \hfill
     \begin{subfigure}[b]{0.22\textwidth}
         \centering
         \includegraphics[width=\textwidth]{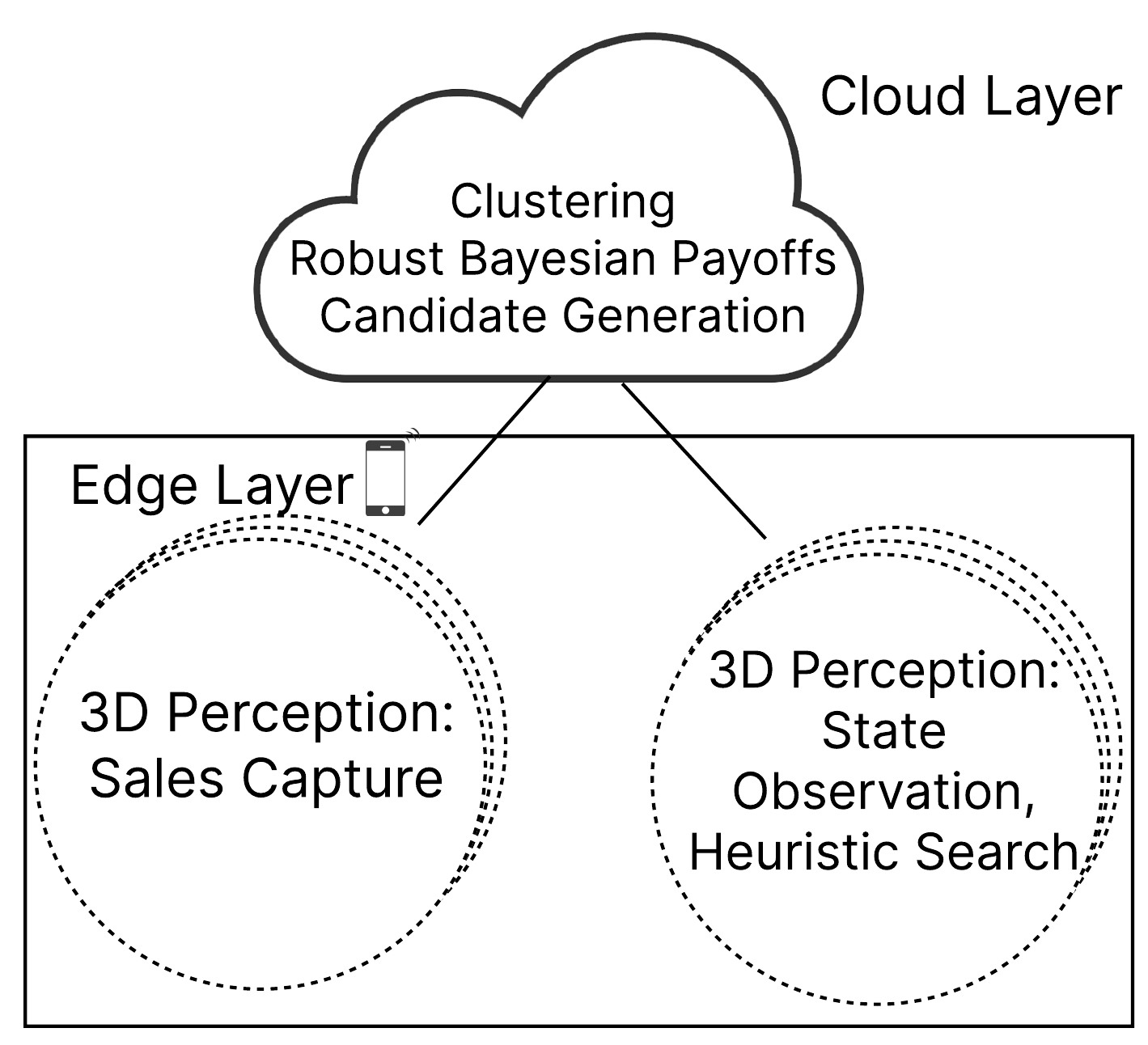}
         \caption{Physical architecture of \ours}
         \label{fig:arch_b}
     \end{subfigure}
     \caption{(a) Logical design of \ours. The user performs successive 3D scans of a display to obtain product count predictions. These predictions are differenced over time following Equation \ref{eq:sales} to estimate sales. We train our two-stage spatial clustering pipeline to group demographically similar stores. Within each cluster, we estimate a Robust Bayesian Payoff (RBP) model to obtain a ranking statistic (PEPF) for each product. We generate candidates for each display, rank each candidate, and produce recommendation sets with heuristic search. (b) Physical design of \ours. 3D perception is implemented at the edge of the network to facilitate real-time feedback. Data are aggregated across nodes for value estimation and candidate generation. When recommendations are served, we observe the state of the display in real-time and apply heuristic search to improve it. }
    \vspace{-5mm}
\end{figure*}

\subsection{3D Data Collection}

The most important input to \ours is fine-grained estimates of product sales over time. Sales estimates need to be both high resolution in time, and specific to each product. Historically, this is done either by hand or by using point-of-sale (POS) data. Hand collected sales data is expensive to collect, while POS is too low resolution to understand how product sales vary by display location within a store. In real-world retail, it is common for a product to be displayed in many locations throughout a single store. Thus, POS data alone does not facilitate targeted recommendations throughout a store.

Instead, we train and deploy CountNet3D \cite{jenkins2023wacv}, a regression-based 3D computer vision model to infer counts of densely spaced objects. CountNet3D takes as input a set of images, and a LiDAR point cloud, and outputs a set of fine-grained, product-level count estimates. This is especially important when the number of available products is in the hundreds or thousands. CountNet3D is only trained to predict object counts in 3D space. In order to collect sales data, CountNet3D is deployed to a mobile device where inference on the images and point cloud data can occur on-edge. The user is a front-line retail worker (i.e. merchandiser), and is the person responsible for restocking the shelves with product, typically on a daily basis. Prior to the restock, the user performs a \textbf{pre-scan} of the shelf or product display. The pre-scan provides a current snapshot of the available inventory on the display. Next, the user restocks the display following a pre-specified plan (called a planogram or schematic). Finally, upon completion of the restock the user performs a \textbf{post-scan}, to capture the inventory when the display has been replenished with product. Using these two inventory snapshots, we can difference the counts over time to estimate sales with the following equation:

{\small
\begin{equation}\label{eq:sales}
    sales_{ijt} = \hat{c}^{(post)}_{ijt} - \hat{c}^{(pre)}_{ij(t-1)}
\end{equation}
}

Where $\hat{c}^{(post)}_{ijt}$, is the post-scan and captures the count estimate for product $i$ at display $j$ in the current timestamp, $t$. Moreover, $\hat{c}^{(pre)}_{ij(t-1)}$ is the prescan in the current visit and captures the remaining inventory at the previous time $t-1$. Intuitively, Equation \ref{eq:sales} captures what product has been removed since the previous merchandiser visit. We assume that all  missing product has been sold. More details about workflow and deployment can be found in the supplement.


\subsection{Spatial Clustering}\label{sec:cluster}

\begin{algorithm}
\small
\caption{SpAGMM Estimation}\label{alg:spagmm}
\begin{algorithmic}
\State \textbf{Input}: Store Coordinates, $S$; Tract Coordinates, $C$; 

\State \textbf{Output}: $K$ x $Z$ Probability Matrix, $\Pi$

\State Initialize Likelihood history, $H = $ []; Number of Clusters, $K$; Number of Tracts, $Z$; Weights, $\pi = \frac{1}{S}I_S$; Covariance Matrix, $\Sigma$ = SPD Matrix; Length of Each Data Vector (a $\langle$ lat, long $\rangle$ pair), $D = 2$

\State{Normalize $S$ to get fixed cluster centers}:
\State $\mu \gets \frac{S - \Bar{S}}{\sigma(S)}$ \Comment{Fix $\mu$. Don't update during training}

\State $\hat{C} \gets \frac{C - \Bar{C}}{\sigma(C)}$

\While{True} \Comment{Estimate the probability that store $s$ belongs to tract $t$:}
    \State $E \sim \mathcal{L}(\hat{C}, \mu, \Sigma)$ \Comment{Expectation Step}
    \State $lik \gets \Sigma_{t}^{Z}\Sigma_{k}^{K}E_{tk}$ \Comment{Get sum of likelihood at step for convergence test}
    \State $H.push(lik)$
    \State $\Pi \gets$ ($K \times Z$) Array
    \For{\texttt{$k \gets 0$ : $K$}} \Comment{Maximization Step}
        \State $\Pi_k \gets \frac{E_k * {\pi}_k}{\Sigma_{k}^K (E_k * \pi_k) + \epsilon}$ \Comment{Get weighted probability of store $s$}
        \State $\pi_k \gets \Bar{\Pi_k}$ \Comment{Update parameters $\pi$ and $\Sigma$ that maximize $E$}
        \State Update $\Sigma$ using MAP estimation to handle CVP:
        \State $Y_0 \gets \frac{1}{S^{1/D}}I_D$
        \State $Y_k \gets \Sigma_{t}^Z \Pi_{tk}(\hat{C_{t}}-S_{k})(\hat{C_{z}}-S_{k})^T$
        \State $r_k \gets \Sigma_{t}\Pi_{tk}$
        \State $\Sigma_k \gets \frac{Y_0 + Y_k}{r_k + 2D + 4}$

    \EndFor
    \State converged $\gets length(H) > 2$ and $\frac{H[-2] - H[-1]}{H[-2]} \leq 0.05$
    \If{converged}
        \State break
    \EndIf
\EndWhile
\State return $\Pi$
\end{algorithmic}
\end{algorithm}

After collecting a time series of product sales data, we segment the data for recommendation using a novel, two-step spatial clustering routine. Intuitively, we want to account for consumer preference heterogeneity using demographic data, since a large body of literature has linked consumer preferences to observable demographic variables \cite{allenby1998marketing, rossi2003bayesian, andersen2010preference}. Consistent with this literature, our expectation is that product preferences vary across these demographic clusters. However, it is typically challenging for retailers to estimate a demographic profile for shoppers as most consumers shop anonymously and do not reveal their demographic attributes. We refer to this as the Demographic Spatial Clustering problem.

To solve this problem we use readily accessible 2020 US census data to create a demographic profile of each store. We are given stores, $s_l \in  \{s_1, s_2, ...\}$, and census areas, or tracts, $\alpha_z \in \{\alpha_1, \alpha_2, ... \} $. The census data consists of $b=30$ demographic variables. Unfortunately, matching stores to census tracts is not obvious. First, we typically have an unequal number of stores and tracts. In rural areas we might have more tracts than stores, and in urban areas we might have more stores than tracts. Second, stores might lie on the boundary of a census tract, making it difficult to simply apply the demographics of the nearest tract.  Third, shoppers from a nearby census tract might visit multiple stores. 

To resolve these issues we propose the Spatially Anchored Gaussian Mixture Model (SpAGMM) to calculate spatially weighted demographic profiles of stores using census tract data. Once a demographic vector of each store is obtained, grouping similar stores together through K-means is straightforward and effective. We detail both steps below.

\subsubsection{Step 1: SpAGMM Estimation}

The core idea behind SpAGMM is we can use a mixture of Gaussians to learn the probability that shoppers from tract $\alpha_z$ visits store $s_l$. This probability is inversely proportional to the distance between $\alpha_z$ and $s_l$. For each store, $s_l$, we can then weight the demographic features of all other tracts, $\alpha_z$, by their respective probabilities to get a spatially weighted, demographic profile. SpAGMM is a special case of the Gaussian Mixture Model (GMM), a common clustering technique used to fit $K$ Gaussians to unlabelled data. Its output is a set of parameters, $\{ \langle \mu$, $\Sigma, \pi \rangle_k \}_{k=1}^{K}$,  which define the cluster mean, covariance, and membership weight. 



We modify the standard GMM by setting the number of clusters equal to the number of stores, $K=L$ and anchoring the Gaussian means at the latitude/longitude coordinates of the stores, $\mu_k = \langle lat, long \rangle_{l}$. Consequently, we do not update $\mu_k$ during training, and only update $\Sigma_k$ and $\pi_k$. We use Maximum A Posteriori (MAP) estimation instead of Maximum Likelihood Estimation (MLE) for numerical stability. When multiple stores and tracts lie in a small neighborhood, the MLE forces the variance of some clusters to be 0, which leads to the Collapsing Variance Problem (CVP) \cite{murphy2012machine}. To tackle this problem, we put a Normal/Inverse Wishart (NIW) prior over $\mu_k$ and $\Sigma_k$ to derive the NIW conjugate model. We use the Expectation-Maximization (EM) algorithm to learn $\Sigma_k$, and $\pi_k$. We provide a formal definition in Algorithm \ref{alg:spagmm}.




Given the outputs of SpAGMM, we can calculate a demographic profile for each store by weighting each tract's demographics by the probability it belongs to $s_l$'s cluster , $X_l = \sum_{z=1}^{Z} \pi_{lz} x_z$

\subsubsection{Step 2: K-means}
Step 1 outputs a single demographic vector, $X_l$ for each store. To model preference heterogeneity, we cluster stores into $K$ groups. We use standard K-means to group similar stores together and provide a visualization of the learned clusters in Figure \ref{fig:cluster} in the supplement.

\subsection{Value estimation: Robust Bayesian Payoffs (RBP)}

After deriving the $K$ clusters, our system learns the posterior predictive distribution of payoff for each product, within each cluster using a Robust Bayesian Payoff (RBP) model. Since the use of a computer vision system introduces noise into the sales data, it is important to develop a technique that can make effective decisions under such observation uncertainty, and therefore avoid costly errors. In physical retail, errors are costly as there is significant shipping and labor cost to merchandise each product. 

RBP is a hierarchical Bayesian model, and has the following advantages over non-Bayesian reward models (i.e., linear regression or deep learning): 1) RBP is more robust to outliers through the use of regularizing priors. 2) RBP is able to pool information across stores within a cluster, but is also able to discount irrelevant data as more data is acquired within a store. This is due to the hierarchical structure of the model, where we learn store and cluster-level parameters. 3) RBP explicitly quantifies uncertainty. In Figure \ref{fig:toy-exp} of the supplement we analyze RBP on these three dimensions with a synthetic dataset. The RBP model can be specified as follows:

\begin{flalign}\label{eq:moments}
\begin{aligned}
    \beta_{i,k} \sim \textrm{HalfNormal}(\mu_0, \sigma_0), 
    b_i \sim \textrm{HalfNormal}(\mu_1, \sigma_1) \\
    \beta_{i,l, k} \sim \textrm{Laplace}(\beta_{i, k}, b_i), 
    \sigma^{(i)}_r \sim \textrm{HalfNormal}(\mu_2, \sigma_2) \\
    r_{i,l,k} \sim \Gamma(x^{(i)}_{q} * \beta_{i,l, k}, \ \sigma^{(i)}_r )
\end{aligned}
\end{flalign}

where $i$ indexes product, $l$ indexes store, and $k$ indexes cluster. $\beta_{i,k}$ and $\beta_{i,l.k}$ denote the cluster- and store-level coefficients for product $p_i$. Moreover, $x^{(i)}_{q}$, is the allocated quantity of product, $p_i$. Thus, the $\beta$ parameters describe how reward changes as a function allocated space. We treat $\{\mu_0, \sigma_0\}, \{\mu_1, \sigma_1\}$, and $\{\mu_2, \sigma_2\}$ as hyperparameters. Finally, $r_{i,l,k}$ is the observed payoff (reward) and is modelled with a Gamma likelihood, which has positive support ($r>0$), since product sales are non-negative. We use HalfNormal priors to ensure that the upper-level coefficients are non-negative, and a Laplace prior to encourage robustness to outliers. We visualize the model with plate notation in Figure \ref{fig:rbp-dag} of the supplement. Since RBP is a fully probabilistic model its output is the posterior predictive distribution, $p(r_{i,l,k} | x^{(i)}_{q})$, or our belief about reward given the amount of space allocated to product, $p_i$. We train RBP using PyMC \cite{salvatier2016probabilistic} and the No U-Turn Sampler \cite{hoffman2014no}. 

Intuitively, we can use $p(r_{i,l,k} | x^{(i)}_{q})$ to derive an uncertainty penalized ranking statistic, Penalized Expected Payoff per Facing (PEPF):

\begin{equation}
    \textrm{PEPF} =  \mathbb{E}[r_{i,l,k}] - \lambda  \sigma(r_{i,l,k})
    \label{eq:epf}
\end{equation}

Where $\mathbb{E}[r_{i,l,k}]$ and $\sigma(r_{i,l,k})$ are the mean and standard deviation of $p(r_{i,l,k} | x^{(i)}_{q})$. We treat the penalty coefficient, $\lambda$, as a user-chosen hyperparameter. Larger $\lambda$ values correspond to more aggressive discounting of uncertain product recommendations. Equation \ref{eq:epf} takes a similar form to Upper Confidence Bound (UCB) estimators in multi-armed bandit problems. UCB estimators \cite{auer2002finite} have been shown to optimally manage the exploration-exploitation tradeoff. However, they systematically take actions with high expected payoff and \textit{high uncertainty}. UCB policies take actions that might increase short-term regret, but obtain more information about an action's payoff and therefore minimize long-term regret. Conversely, recent work has shown that using an uncertainty penalized reward estimator can encourage safe policies \cite{yu2020mopo, jenkins2022bayesian} in real-world reinforcement learning. In this penalized regime, the policy tends to choose actions with high expected payoff and \textit{low uncertainty}. Additionally, Yu et al. \cite{yu2020mopo} prove that the penalized reward estimator maximizes the lower bound of the true reward. We prefer the uncertainty penalized estimator over UCB because it is both theoretically sound and useful in real-world problems where errors are costly. In our experiments, we set $\lambda=1$.

\subsection{Candidate Generation}

We generate candidates to reduce the search space of products. We desire products that are physically and semantically compatible with the observed state of the display. Our candidate generation is comprised of three major steps: graph construction, sampling, and pruning. The initial product set observed on each display is treated as a ``seed set'' to seed the candidate generator.



\textbf{Graph Construction:} Mining a live database of observed display states (a list of products on a display at a specific time) we construct a weighted graph of products. Each node is a product and each edge is the number of times that two products appear together on the same display. Products that appear together frequently have very large edge weights while products that appear together rarely have small edge weights. Each product in our dataset is assigned to one of 5 sub-categories (Sparkling, Water, Isotonic, Rejuvenate, or Energy). We restrict our graph to only contain edges between products of the same sub-category. This graph partitioning ensures product consistency between the candidate set and the seed set. The graph is updated on a weekly basis.


\textbf{Sampling:} 
To create the candidate set we first sample from neighbors of the seed set, proportional to the edge weights. For each seed product $p_i$, we sample $\tau$ products without replacement from its neighbors with respect to the edge weights. Each product sampled is given one “vote”, which are then summed and normalized across all of the seed products. The output of this operation is an initial candidate set. 



\textbf{Pruning:} 
We prune the set of candidates based on known product dimensions from a product database. We compare the maximum height of any product in the “seed” set against the initial candidate set, and remove any products whose height exceeds that of the max of the seed set. This step ensures candidates will physically fit on the product display.

\subsection{Heuristic Search}


\begin{algorithm}
\caption{Heuristic Search}\label{alg:search}
\begin{algorithmic}
\small
\State \textbf{Input}: Candidate set, $\mathcal{C}$, Observed state set $\mathcal{S}$, scoring function $f(\cdot)$, decay function $\delta(\cdot)$, number of products to swap $V$, time steps $T$, epsilon-greedy parameter $\epsilon$, maximum product facings $M$, product-quantity lookup $Q_s$; 
\State \textbf{Output}: Recommended product-quantity lookup $R$
\State $B \gets M$, assign budget, $B$
\State $R \gets$ KeyValue[], Initialize recommendation lookup
\For{$t$ in $T$}
    \State $\mathcal{C} = f(\mathcal{C})$ \Comment{score candidates, assign PEPF}
    \State $\mathcal{S} = f(\mathcal{S})$ \Comment{score current state, assign PEPF}
    \State $\Tilde{\mathcal{C}} = \textrm{sort}(\mathcal{C})$ \Comment{sort candidates by PEPF ascending}
    \State $\Tilde{\mathcal{S}} = \textrm{sort}(\mathcal{S})$ \Comment{sort candidates by PEPF ascending}
    \While{$B > 0$}
        \For{$j$ in $V$} \Comment{search for new products}
            \State $p \gets \mathcal{S}[0]$ \Comment{select lowest PEPF product}
            \State $q \gets Q_s[p]$ \Comment{current quantity}
            \State $\alpha \sim \textrm{Unif}(0, 1)$ \Comment{$\epsilon$-greedy action}
            \If{$\alpha < \epsilon$}
                \State $p' \sim \Tilde{\mathcal{C}}$ \Comment{choose random candidate}
            \Else
                \State $p \gets \Tilde{\mathcal{C}}$.pop() \Comment{greedy selection (highest PEPF)}
            \EndIf
            \State $q' = \delta(q)$ \Comment{decay quantity}
            \State $R[p'] \gets q'$ \Comment{assign recommendation}
            \State $B = B - q'$
            \State $Q_s[p] = q - q'$ \Comment{update product quantity}
        \EndFor
        \State $p' = \Tilde{\mathcal{S}}$.pop() \Comment{Collect existing, high PEPF products}
        \State $q' = Q_s[p']$
        \State $R[p'] \gets q'$
        \State $B = B - q'$
    \EndWhile
\EndFor
\end{algorithmic}
\end{algorithm}

Using the candidates as input, we next perform a heuristic, combinatiorial search to produce a recommended product set.  Due to the high stakes nature of physical world recommendation systems, we design a heuristic search algorithm that is relatively simple, in order to avoid catastrophic failures and lost sales. The core principle behind the search algorithm is to keep high payoff products, and slowly trade low payoff products for better ones, while accounting for uncertainty.

Our search algorithm is comprised of the following general steps. First, we produce a PEPF score (Equation \ref{eq:epf}) for each product in the candidate set, $\mathcal{C}$, and the observed  state set, $\mathcal{S}$. Second, we sort the candidate set and state set by PEPF. Third, we choose the lowest $V$ products by PEPF and reduce their quantities following the decay schedule $\delta(t, q_0) = \lfloor q_0\frac{1}{2}^t \rfloor$, where $q_0$ is the first observed product quantity and $t$ is the time step, or the $t^{th}$ time the product has been selected. The floor term ensures that the output is a discrete count, and that eventually, the decay will drop to zero. This decay schedule will slowly remove poorly performing products until they are completely removed.  Fourth, we re-allocate the removed space with new products following an $\epsilon$-greedy routine. With probability $\epsilon$, we choose a random product from the candidate set, and with probability $1-\epsilon$ we choose a greedy action, or the candidate with the highest PEPF. Fifth, after we have $V$ product swaps, we fill the remaining budget with the high PEPF products already in the state, $\mathcal{S}$. We formally describe our search in Algorithm \ref{alg:search}.

\section{Experiments}\label{sec:exp}

In the following section we discuss our experimental setup to validate that \ours is an effective, real-world recommendation system for shelf space allocation. We perform both offline and online evaluation. Notably, we partner with a large beverage distribution company and perform two real-world experiments over the course of 2022. We then deploy the system to a larger set of users and stores in 2023. 


\subsection{Offline Evaluation}\label{sec:offline}

We perform an offline comparison of \ours to search algorithms that have been applied to the product assortment problem in prior work. Offline policy evaluation typical of reinforcement learning and bandit algorithms is notoriously difficult in part due to the `partial label' problem \cite{li2010contextual, li2011unbiased, castells2022offline, mandel2016offline}. We apply the evaluator proposed by Li et al. \cite{li2011unbiased}. To build a static dataset, we collect user logs from the $\textit{control}$ group in experiment \#2.  This dataset is comprised of 33 stores, and covers approximately 8 weeks of time, and includes 14,830 transactions. It is important to note that the control data is a fair comparison because no online intervention is ever made by any of these search algorithms. The evaluator in \cite{li2011unbiased} assumes the logged actions are \textit{iid}. We randomly sample 50\% of our logged interactions to reduce temporal dependency in the dataset.

We compare \ours to a variety of classic and modern baselines. We run each algorithm 30 times, and report the mean, standard deviation (in parentheses) and median payoff in Table \ref{tab:offline}. Both Deep Ensembles \cite{lakshminarayanan2017simple} and Model-based Offline Policy Optimization (MOPO) \cite{yu2020mopo} feature modern neural networks with predictive uncertainty. For Deep Ensembles, we use the same heuristic search as \ours to measure the value of RBP. MOPO uses uncertainty penalized offline reinforcement learning for action selection. We observe that \ours outperforms all existing baselines in both mean and median payoff, indicating it can effectively take good actions under uncertainty.

Additionally, we perform an ablation study of \ours using the same dataset and experimental setup described above. The results are reported in Table \ref{tab:ablation}. We carefully remove each of the core components: clustering, Robust Bayesian Payoffs (RBP) and heuristic search. When we remove clustering, we fit a single, global reward model. When we remove RBP, the reward model is a non-Bayesian, simple linear regression. When we remove the heuristic search module, we apply a greedy search without the penalized reward ranking statistic. Overall, the heuristic search and Bayesian reward model have a large impact on performance and \ours outperforms all ablated variants.

{\small
\begin{table}[]
    \centering
    \begin{tabular}{ccc}  
        \toprule
         Algorithm & Mean Reward $\uparrow$ & Median Reward $\uparrow$ \\
         \toprule
         Dynamic Program & 1.001 (1.8) & 0.0   \\
         Linear Program  & 0.718 (1.6) & 0.0  \\
         $\epsilon$-greedy  & 1.790 (1.6) & 1.864  \\
         Genetic    & 2.147 (1.8) & 2.265  \\
         Deep Ensembles & 1.895 (1.6) & 2.452 \\
         MOPO & 1.826 (1.6) & 2.048 \\
         \ours   & \textbf{2.466} (1.6) & \textbf{2.496}  \\
         
    \end{tabular}
    \caption{Offline Evaluation using a static dataset and the offline policy evaluator algorithm from \cite{li2011unbiased}}.
    \label{tab:offline}
    \vspace{-10mm}
\end{table}
}

\subsection{Online Experimental Phase}

\begin{table*}[]
    \small
    \centering
    \begin{tabular}{cccccccccc}
          \toprule
          & Dates & Pre/Post (Days) & \# Stores & \# Displays &  Group & Avg Reco. Compliance  & DID & $p$-val & DID IQR  \\
         \toprule
        \multirow{3}{*}{Exp. 1} & \multirow{3}{*}{5/17 - 6/30} & \multirow{3}{*}{25/20} & \multirow{3}{*}{2} & \multirow{3}{*}{22} &  Store \#1 & 100\%  & $+35.10\%^{\ast\ast}$ & 0.014 & [23.65\%, 46.08\%] \\
         &   &  & &  & Store \#2 & 100\%  & $+32.04\%^{\ast}$ & 0.064 & [15.24\%, 44.27\%] \\
         &   & & & &  All &  100\%   & $+35.03\%^{\ast\ast}$ & 0.016  & [23.42\%, 42.78\%] \\
        \midrule
         \multirow{2}{*}{Exp. 2} & \multirow{2}{*}{9/21 - 11/19} & \multirow{2}{*}{21/38} & \multirow{2}{*}{33} & \multirow{2}{*}{133} &   All & 61.64\%  & +07.83\% & 0.222 & [1.67\%, 13.65\%] \\
         &  &  & & &  High & 90.38\%  & +27.78\%$^{\ast \ast}$& 0.003 & [21.12\%, 35.18\%] \\
    \end{tabular}
    \caption{Overall summary of our two field experiments based on the Difference-in-Difference (DID) experimental design. Where recommended changes are adopted (high compliance), we see large increases in sales that are statistically meaningful. We use $^{\ast \ast}$ to denote significance at $\alpha = 0.05$, and  $^{\ast}$ for significance at $\alpha = 0.1$.}
    \vspace{-7mm}
    \label{tab:results}

\end{table*}

\subsubsection{Experimental Design}\label{sec:exp-design}

\begin{table}[]
    \centering
    \begin{tabular}{ccccc}
         \toprule
         Exp. & Clustering & RBP & Heuristic Search & Reward   \\
         \toprule
         \# 1 & \xmark & \xmark & \xmark &  0.866 (1.6)\\ 
         \# 2 & \xmark & \xmark & \cmark &  2.020 (1.4)\\ 
         \# 3 & \cmark & \xmark & \xmark &  0.585 (1.4)\\ 
         \# 4 & \cmark & \xmark & \cmark &  2.143 (1.4) \\
         \# 5 & \xmark & \cmark & \xmark &  0.998 (1.7) \\ 
         \# 6 & \xmark & \cmark & \cmark &  2.308 (1.7) \\
         \# 7 & \cmark & \cmark & \xmark &  0.490 (1.2) \\ 
         \# 8 & \cmark & \cmark & \cmark &  \textbf{2.466 (1.6)}  \\ 
         
    \end{tabular}
    \caption{Offline ablation study using a static dataset and the offline policy evaluator algorithm from \cite{li2011unbiased}}
    \vspace{-10mm}
    \label{tab:ablation}

\end{table}

Controlled experiments in real-world scenarios can be a challenging task due to the possibility of confounding, or omitted variables. To account for potential confounders, we use a \textit{within-store} Difference-in-Difference (DID) experimental design for both of our field experiments. DID estimation is a commonly used experimental design technique in the Econometrics literature to assess causal relationships \cite{angrist2009mostly, dinardo2010natural}. The key idea behind the DID experimental design is to measure the change in a response variable before and after an intervention for both the affected and unaffected groups \cite{bertrand2004much}. The difference of these changes is the treatment effect. More formally, $\widehat{\textrm{DID}} = (\texttt{post}_{\textrm{treat}} - \texttt{pre}_{\textrm{treat}}) - (\texttt{post}_{\textrm{control}} - \texttt{pre}_{\textrm{control}})$, where \texttt{pre} and \texttt{post} are pre-treatment and post-treatment periods respectively. Our controlled experiments have the following design steps:

\begin{itemize}
    \item We mark a set of product displays in each store for observation. A product display is a shelf, cooler, end-rack or other merchandising display. Each display is uniquely identified via QR code.
    \item \textit{Within} each store, we randomly assign each display to either treatment or control with 50\% probability.
    \item The control displays are merchandised according to current industry best practices. Treatment displays are merchandised with the recommended actions from EdgeRec3D.
    \item For both treatment and control, the user (retail worker or merchandiser) visits the display typically on a daily basis. A QR code is scanned to uniquely identify that display. The user replenishes the product, and uses a mobile app to observe the state and record new sales. 
    \item Pre-treatment: Sales are recorded for both treatment and control. For both groups, we monitor sales without making recommendations for a fixed period to get a sales baseline.
    \item Post-treatment: After the pre-treatment period has ended, EdgeRec3D begins serving product recommendations for the treatment group only.
    \item At the conclusion of the post-treatment period, we calculate the difference-in-difference estimator, $\widehat{\textrm{DID}}$.
    
\end{itemize}

Note that recommendations are simply displayed to users (retail workers or merchandisers); ultimately, each user has the autonomy to accept or reject the EdgeRec3D suggestions. Some fail to accept due to time constraints, while others are simply resistant to using the system and prefer existing practice. We measure this acceptance rate with a compliance statistic, the Jaccard index between the recommended state, $R$, and the observed state $\mathcal{S}$: $\rho(R, \mathcal{S}) = J(R, \mathcal{S})$.



In our first experiment, we choose two stores in the Salt Lake City, Utah, USA region and focus the test on cooler displays.  We run a six week experiment, from May 17 to June 30, 2022 using a DID design. The treatment period begins on June 10, 2022. Our second experiment is much larger and consists of 33 stores across the mountain west region of the United States (Utah, Colorado, Arizona, etc...). We again focus on cooler displays. Experiment \#2 runs from September 21, 2022 to November 19, 2022, or approximately eight weeks. The treatment period begins on October 21, 2022. Due to the time and expense of online, controlled experiments we do not compare \ours to the benchmark search algorithms studied in Section \ref{sec:offline}. Instead, during our online evaluation we compare \ours to current industry best practices.

\subsubsection{Results}

We provide an overall summary of our two field experiments in Table \ref{tab:results}. To obtain estimates of statistical confidence we perform a non-parametric permutation test \cite{good2013permutation}. Across the two experiments, we see a +27.78\% and +35.03\% increase in average daily sales where recommendations are followed. These results are both statistically significant ($\alpha=0.05$) and provide strong evidence that implementing and following the recommended product assortment decisions prescribed by \ours have a positive, causal impact on daily average product sales.

\paragraph{Field Experiment \#1}

\begin{figure}
    \centering
    \includegraphics[width=.5\textwidth]{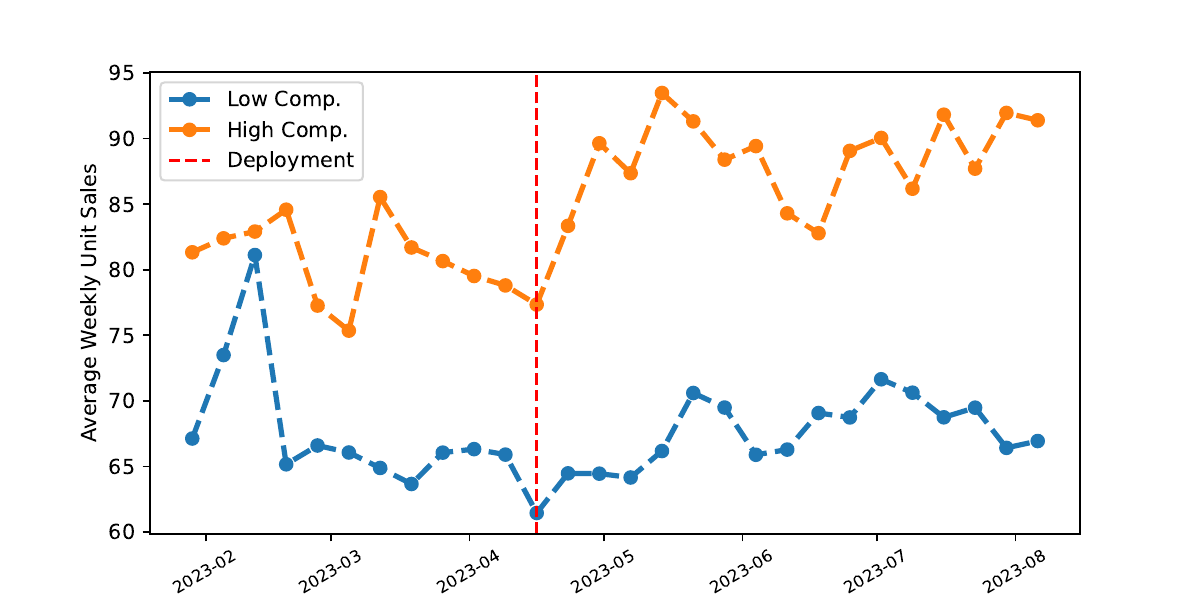}
    \caption{Deployment phase observational study. We split stores into high a low compliance groups as in the experimental phase. We see that the two groups have roughly parallel trends in the pre-deployment period.  We deploy \ours on April 16, 2023. In the post-deployment period, the two groups are still subject to the same seasonal trends, but the high compliance group diverges from the low compliance group. See Table \ref{tab:deployment} for full results.}
    \label{fig:deployment}
    \vspace{-5mm}
\end{figure}

In the first experiment, we deploy \ours to only two stores with expert users. These users are able to take their time, and the implementation of recommendations served by \ours is strictly enforced. Therefore, during experiment \#1, the average recommendation compliance is 100\%. We see a very high DID treatment effect of +35.03\%. 

Daily product sales in the treatment group increased by 0.95 units (+16.42\%), while the control group saw a decrease of 1.0 units per day (-18.61\%). The difference between these two yields the total treatment effect, $16.42\% - (-18.62\%) =  35.03\%$.  The control group offers a counterfactual estimate of what would have happened in the absence of the intervention (\ours). This is known as the parallel trend assumption of DID \cite{angrist2009mostly}. Had we not deployed recommendations, it's likely that average daily sales would have decreased at similar rates as the control group. 



\paragraph{Field Experiment \#2}

 In this experiment, our user base consists of non-experts who have more time constraints (ie, other retail responsibilities), which is reflective of the real-world product assortment problem. Here, the user has the option to adopt  the recommendation from \ours. Consequently, we see more variation in recommendation compliance during experiment \#2. 

The DID treatment effect across all stores is +7.83\%. While this result is positive, our confidence is not high enough to conclude statistical significance ($p$-value = 0.222). However, the average recommendation compliance across all stores is only 61.64\%. This suggests that a large number of users are ignoring the recommendations served by \ours. Filtering to stores where recommended product sets are adopted (average compliance $\geq 80\%$), shows a different result. Among this high compliance group we see very strong, positive results where the DID treatment effect is +27.78\% and is significant at $\alpha=0.05$. This leads us to conclude that high recommendation compliance leads to a large treatment effect. This aligns with our intuition because our system can only make recommendations, but it ultimately relies on a user to adopt and implement the change in the physical world. This relationship between compliance and treatment effect across our two experiments is captured in Figure \ref{fig:compliance} of the supplement.

\vspace{-3mm}

\subsection{Deployment Phase}

Following our two controlled experiments, we deploy \ours  to 83 stores and monitor the impact on store-level sales for a period of 28 weeks (11 weeks pre-deployment and 17 week post-deployment). We again use a DID study design. However, this deployment phase differs from Experiment \#1 and \#2 in the following ways. First, recommendations are applied to all cooler displays; we do not randomly assign displays to treatment and control. Instead of a controlled experiment, we perform an observational study to monitor the effectiveness. Second, while \ours still uses 3D perception to estimate sales, here we measure performance using syndicated point-of-sale (POS) data. POS data allows us to observe sales prior to the deployment of our system, and quantify the impact of \ours post-deployment. The data contains sales information for all products and all locations within the store. We filter the POS data to products observed and tracked by \ours in our study. We are not able to disambiguate sale location in this data. Third, the syndicated POS data is reported weekly, instead of daily.

Results of our deployment are reported in Table \ref{tab:deployment} and visually depicted in Figure \ref{fig:deployment}. Since we no longer have treatment and control groups in our deployment phase, we use recommendation compliance as a measure of adoption. We again split stores into high and low compliance groups by calculating the average compliance over the course of the 17 week deployment period. Crucially, we also monitor average sales for these two groups \textit{prior} to deployment to get a baseline performance measurement.

 High compliance stores averaged sales of 80.91 units sales prior to the deployment date, and increased to an average of 88.01 post-deployment date, a change of 7.1 units or 8.8\%. In contrast, low compliance stores averaged sales of 67.80 prior to the deployment date and stayed roughly the same, decreasing slightly, to 67.37 post-deployment date. These figures indicate a 7.53 unit per-product, per-week “treatment” effect on sales for the high compliance stores, which represents a 9.4\% increase in weekly sales volume. This treatment effect is statistically meaningful ($p=0.021$).

 The results in the deployment phase are positive and meaningful, but lower in magnitude than the experimental phase. This is likely explained by the use of controlled experiments, which enable precise estimates of the treatment effect. In spite of this, we still see positive and statistically meaningful results as \ours is deployed.

\begin{table}[]
    \centering
    \begin{tabular}{ccc|c}
         \toprule
         &  \textbf{Low Comp.} & \textbf{High Comp.} & \textbf{Diff-in-Diff}  \\
         \toprule
         Pre & 67.80 & 80.91 &  \\ 
         Post & 67.37 & 88.01 &   \\ 
         \midrule
          Change  &  -0.43 (-0.6\%)  & +7.1 (+8.8\%) & $\textbf{+7.53 (+9.4\%)}^{\ast\ast}$\\
    \end{tabular}
    \caption{Observational study during the deployment phase of \ours. We show weekly average units sold before and after the intervention (\ours). The high compliance group increases by 7.1 (8.8\%) weekly average units, while the low compliance group stays roughly the same. We calculate the percentage changes and difference these estimates to get the DID treatment effect of 9.4\%. This estimate is statistically significant with $p=0.021$}
    \vspace{-11mm}
    \label{tab:deployment}

\end{table}

\section{Related Work}\label{sec:related}

\textit{Shelf Space Allocation} The shelf space allocation problem has been studied for many years in the Operations Research field \cite{curhan1973shelf}. Classic optimization methods are commonly applied such as linear \cite{yang1999study, geismar2015maximizing}, or dynamic \cite{zufryden} programming, and game theory \cite{martinez2011competing}. More computational methods such as genetic algorithms \cite{ZHENG2023101251, castelli2014genetic, urban1998inventory, hwang2005model} and simulated annealing approaches \cite{borin1994model} have been proposed in recent years. Others have sought to jointly optimize both product assortment and price \cite{murray2010joint} via a branch-and-bound algorithm. \cite{dusterhoft2020practical} performed a controlled field test and showed a 9\% increase in profit using a non-linear program to choose product assortment.

\textit{Offline Recommender Systems} The majority of work in recommender systems has been done online, in the e-commerce domain, but systems for offline commerce are growing in popularity \cite{colombo2020recommender}. Early work \cite{pathak2010empirical, walter2012moving}  showed that recommender systems can positively impact offline product demand. Later work focused on collaborative filtering \cite{sun2014big, park2019group} and association rule mining \cite{tatiana2018market, pandit2010intelligent, chen2007data, brijs1999using}. Machine Learning techniques such as gradient-boosted trees \cite{silva2020recommender}, K-nearest neighbors \cite{leininger2020advancing}, ensemble techniques \cite{bae2010integration} and offline Reinforcement Learning \cite{jenkins2022bayesian} have been studied. Location-based systems leverage customer location data from smartphones \cite{chatzidimitris2020location} or RFID tags \cite{chen2015real} within a store. In contrast, \ours provides real-time assortment recommendations based on fine-grained sales estimates obtained via 3D computer vision, while also accounting for measurement error and preference heterogeneity across stores.

\section{Conclusion}\label{sec:conclusion}

We proposed \ours  to solve the product assortment problem in physical retail. \ours uses an edge-first architecture with real-time 3D perception to estimate product sales and observe display states. Making recommendations at the edge of the network closes the feedback loop and facilitates faster preference discovery. Additionally, \ours relies on a probabilistic reward model (RBP) for uncertainty quantification, spatial clustering to account for preference heterogeneity, and candidate generation with heuristic search to tackle combinatoric explosion. Controlled experiments and a 28 week observational study showed that \ours has meaningful, real-world impact.


\bibliographystyle{ACM-Reference-Format}
\bibliography{ref}

\newpage
\onecolumn


\twocolumn

\section{Appendix}

In what follows we provide a technical appendix to accompany the main body of the paper. This appendix features two main sections. First, we provide additional methodological Details. Second, we present additional experimental results to further strengthen the claims made in the paper.

\subsection{Table of Notation Used}\label{sec:notation}

In this section we provide three tables describing the notation used throughout the paper. Each table corresponds to a different sub-section of the main paper, namely Problem Definition (Table \ref{tab:notation-prob}), SpAGMM (Table \ref{tab:notation-spagmm}), and Heuristic Search (Table \ref{tab:notation-search}).

\begin{table}[]
    \centering
    \begin{tabular}{c|c}
        \toprule
         Variable & Quantity \\
         \toprule
         $N$ & number of products \\
         $p_i$ & product, indexed by $i$ \\
         $q_i$ & the quantity of product $i$ \\
         $d_j$ & the product display index by $j$ \\
         $s_l$ & store indexed by $l$ \\
         $M_j^{l}$ & the maximum capacity of display $j$ in store $l$ \\
         $r_i$ & Observed reward for product $i$\\
         $t$ & time of observation \\
         $a_z$ & spatial areas (census tracts) indexed by $z$. \\
         $x_z$ & vector of demographic statistics for a spatial area, $z$\\
         $b$ & number of demographic features \\
         $K$ & the number of store clusters \\
         $Z$ & number of spatial areas \\
         $\hat{c}_{ijt}$ & estimated counts of $p_i$ in $d_j$ at time $t$ \\
         $\beta_{i, k}$ & reward coefficient for product $i$ in cluster $k$ \\
         $\beta_{i,l,k}$ & reward coefficient for product $i$, store $l$, in cluster $k$ \\
         $\sigma_{r}$ & variance of reward (prior) \\
         $\rho$ & compliance statistic (distance function over sets) \\
         
    \end{tabular}
    \caption{Notation Summary of Problem Definition}
    \label{tab:notation-prob}
\end{table}

\begin{table}[]
    \centering
    \begin{tabular}{c|c}
        \toprule
         Variable & Quantity \\
         \toprule
         $S$ & store coordinates \\
         $C$ & tract coordinates \\
         $H$ & history of likelihood values \\
         $\pi$ & mixture probabilities \\
         $\mu_k$ & mean of $k^{th}$ Gaussian (fixed) \\
         $\Sigma_k$ & covariance matrix of the $k^{th}$ Gaussian  \\
         
    \end{tabular}
    \caption{Notation Summary of SpAGMM}
    \label{tab:notation-spagmm}
\end{table}

\begin{table}[]
    \centering
    \begin{tabular}{c|c}
        \toprule
         Variable & Quantity \\
         \toprule
         $\mathcal{C}$ & set of product candidates \\
         $\mathcal{S}$ & set of observed products (state) \\
         $\mathcal{R}$ & set of recommended products \\
         $f(\cdot)$ & scoring function \\
         $V$ & number of product swaps per iteration \\
         $T$ & number of iterations, or time steps \\
         $\epsilon$ & epsilon-greedy parameter \\
         $Q_s$ & lookup table expressing quantity of each product \\
         $B$ & quantity budget \\
         $\delta(\cdot)$ & decay function \\

    \end{tabular}
    \caption{Notation Summary of Heuristic Search}
    \label{tab:notation-search}
\end{table}

 \subsection{SpAGMM Estimation} We present the full pseudo-code in Algorithm \ref{alg:spagmm}. We use the Expectation-Maximization (EM) algorithm to learn the SpAGMM clusters. In the expectation step, we assign each tract to the nearest cluster given the current estimates of covariance. The maximization step, performs two functions. First, it returns a weighted probability matrix, $\Pi$, defining the probability store $s_l$ matches to tract $\alpha_z$, given $\pi$. Second, it updates the parameters $\Sigma$ and $\pi$ that maximize $E$ returned from the expectation step. We use Maximum A Posteriori (MAP) estimation instead of Maximum Likelihood estimation (ML) due to numerical stability. The MLE is likely to overfit, leading to the variance parameter to collapse to 0.



 \subsection{\ours parameter settings} During our experiments we set the number of product swaps per iteration, $V=2$. At each iteration, we take the lowest $V=2$ products by PEPF and choose 2 higher value product swaps, following an $\epsilon$-greedy routine. We set the $\epsilon$-greedy parameter to $\epsilon=0.05$. We allow for a two week pre-treatment period to get a baseline for sales, following which we deploy recommendations. We update recommendations on a weekly cadence. Finally, we set the uncertainty coefficient $\lambda=1.0$

\subsection{Complexity Analysis of \ours}

Equation \ref{eq:factorial} of the Introduction describes the number of assortment possibilities for $p$ products and $m$ discrete product locations. Since this equation grows factorially in both $p$ and $m$, exploring all assortment possibilities is impossible in finite time for realistic $m$ and $p$. We design \ours to address this combinatorial explosion through two of its components: 1) candidate generation and 2) heuristic search.

\subsubsection{Complexity of Candidate Generator} The candidate generator mines an active database to produce a graph representing the likelihood of product co-occurrences. We then sample from this graph using a ``seed'' set of products, which are the products observed on the current display state. The complexity of candidate generation is

\begin{equation}
    O(n + sp\tau + \tau)
\end{equation}

for $n$ user logs, $s$ size of seed set $\tau$ number of samples from the graph. The algorithm steps through each of the $n$ logs and increments a counter of pairwise-product co-occurence to produce an adjacency list representing the graph. Once this is complete, we sample the graph to produce $c$ candidates. In the sampling step, for each of the $s$ products we sample from it's $p$ neighbors $\tau$ times. In the worst case, the number of adjacent products is $p$ although it practice it is usually much less. We then step through each of the $\tau$ candidates in the pruning step.

\subsubsection{Complexity of Heuristic Search} The heuristic search module is run each time a new recommendation is generated. It's complexity is:

\begin{equation}
    O(c + s + mv)
\end{equation}

for $c$ candidates, $m$ product slots and $v$ product swaps. We first score each of the $c$ candidates and $s$ seed products with the RBP value function to produce a PEPF ranking statistic. Upon completion, for each of the available $m$ product slots, we search for $v$ superior product swaps.

On the whole, \ours is at worst linear in the number of $p$ products and $m$ product slots. From a complexity perspective, this is far more efficient than the factorial growth of a brute force approach described in Equation \ref{eq:factorial}.

\begin{figure}
    \centering
    \includegraphics[width=0.6\linewidth]{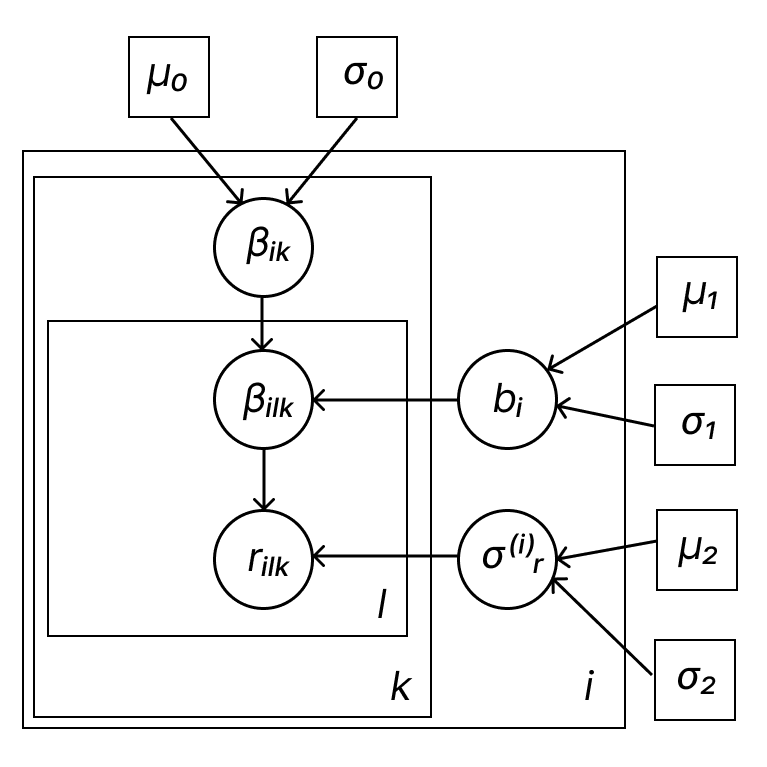}
    \caption{Representation of the Robust Bayesian Payoff (RBP) model using plate notation. $i$ indexes product, $k$ indexes cluster, and $l$ indexes store. Circles indicate random variables and squares indicate hyperparameters. The $\beta$ parameters describe how expected payoff changes as allocated product quantity increases and have hierarchical structure. This hierarchy facilitates store- and cluster-level preference estimation. The variance parameters are not expected to vary across stores or clusters and are estimated universally for each product.}
    \label{fig:rbp-dag}
\end{figure}

\subsection{Wall-clock time of \ours}

Figure 2(b) provides a physical architecture diagram of EdgeRec3D. The components that run on the edge layer (i.e., mobile device) return results in real-time. Specifically, the sales capture and state observation are computed with CountNet3D on the Apple Neural Engine and return results in ~200ms. Additionally, once the candidates are queried from a remote server, the heuristic search algorithm is run in real-time and returns results in <100ms. 

\subsection{Training of Visual Layer}

The visual perception layer of \ours is built on top of CountNet3D \cite{jenkins2023wacv}, a regression-based neural network that processes images and LiDAR scans to predict finegrained object counts. CountNet3D uses a mature 2D object detector, such as YOLOv5 \cite{glenn_jocher_2022_7347926} to propose 3D regions around known objects. It then uses a PointNet-style architecture \cite{qi2017pointnet} to count objects to process the point cloud within each region.

\begin{figure*}

    \begin{tabular}{ccc}
     \includegraphics[width=0.28\textwidth]{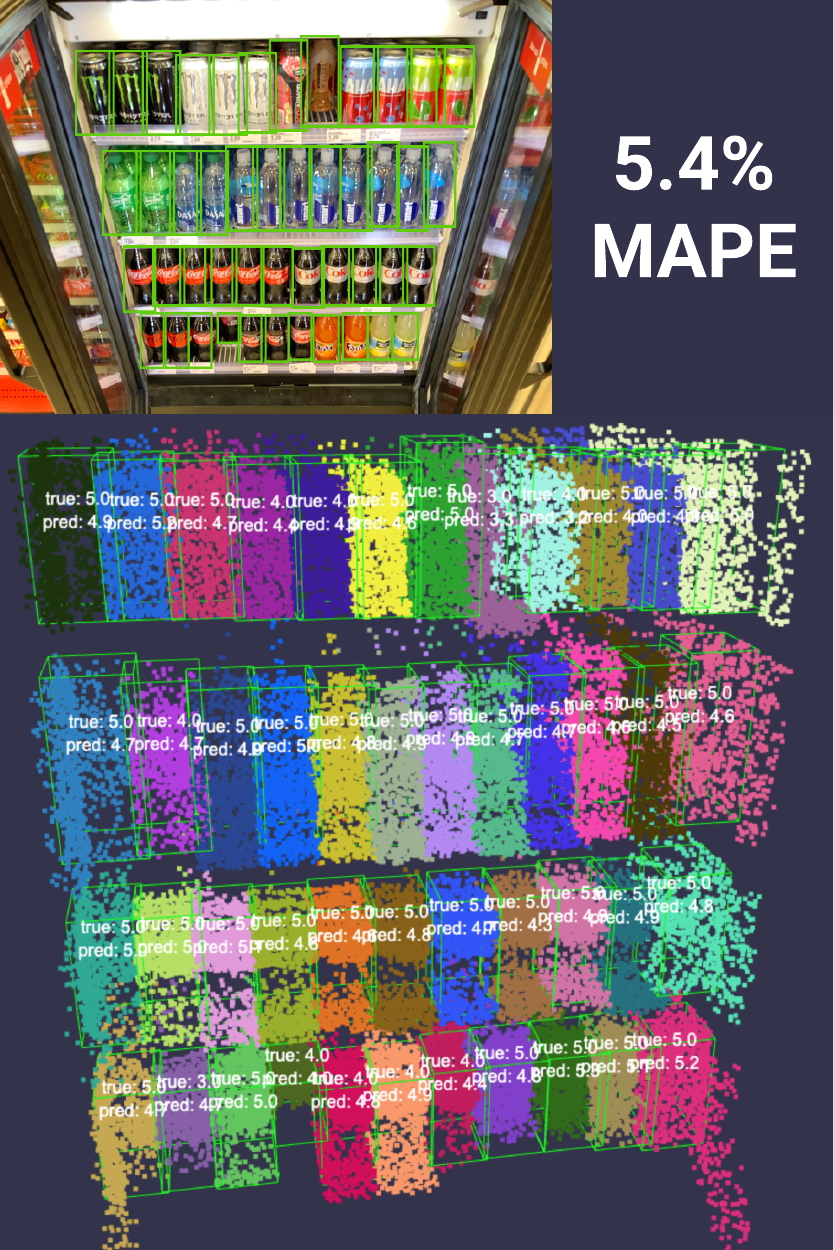} & 
     \includegraphics[width=0.28\textwidth]{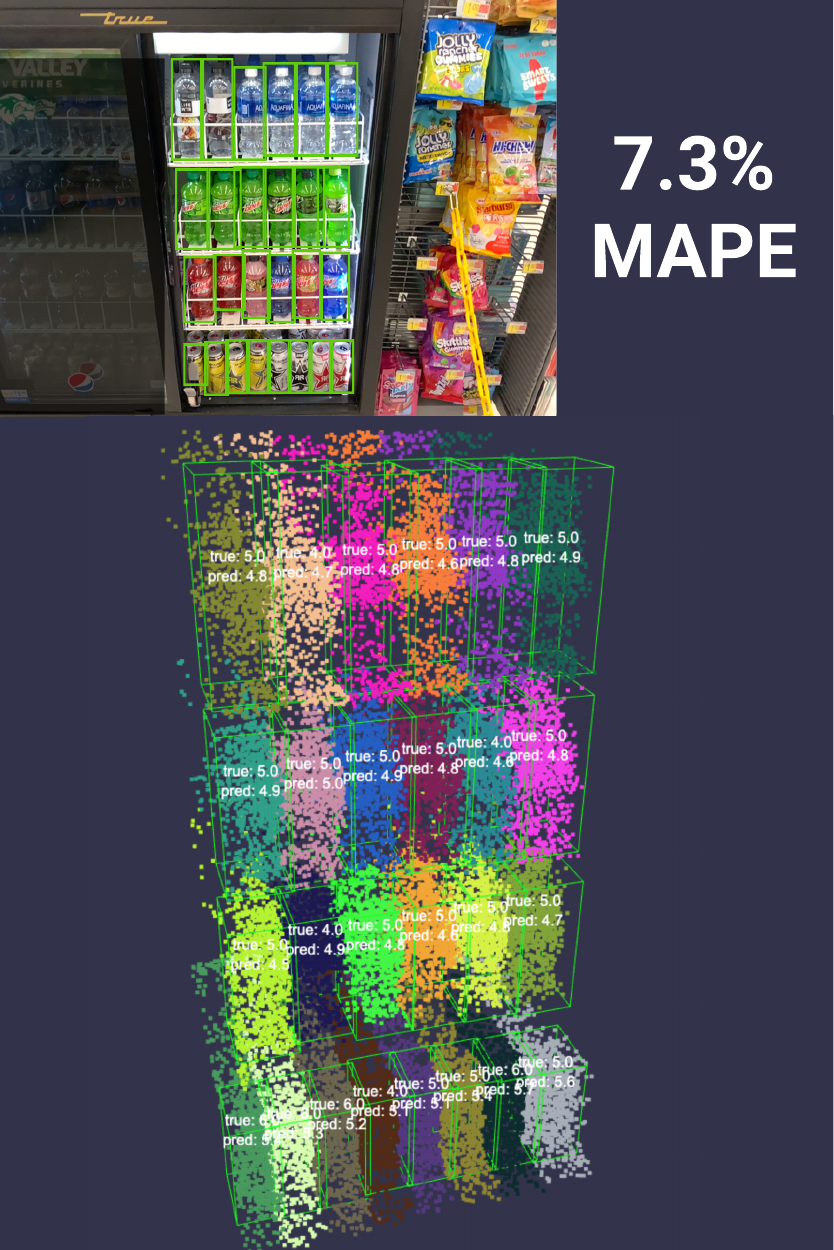} &
     \includegraphics[width=0.28\textwidth]{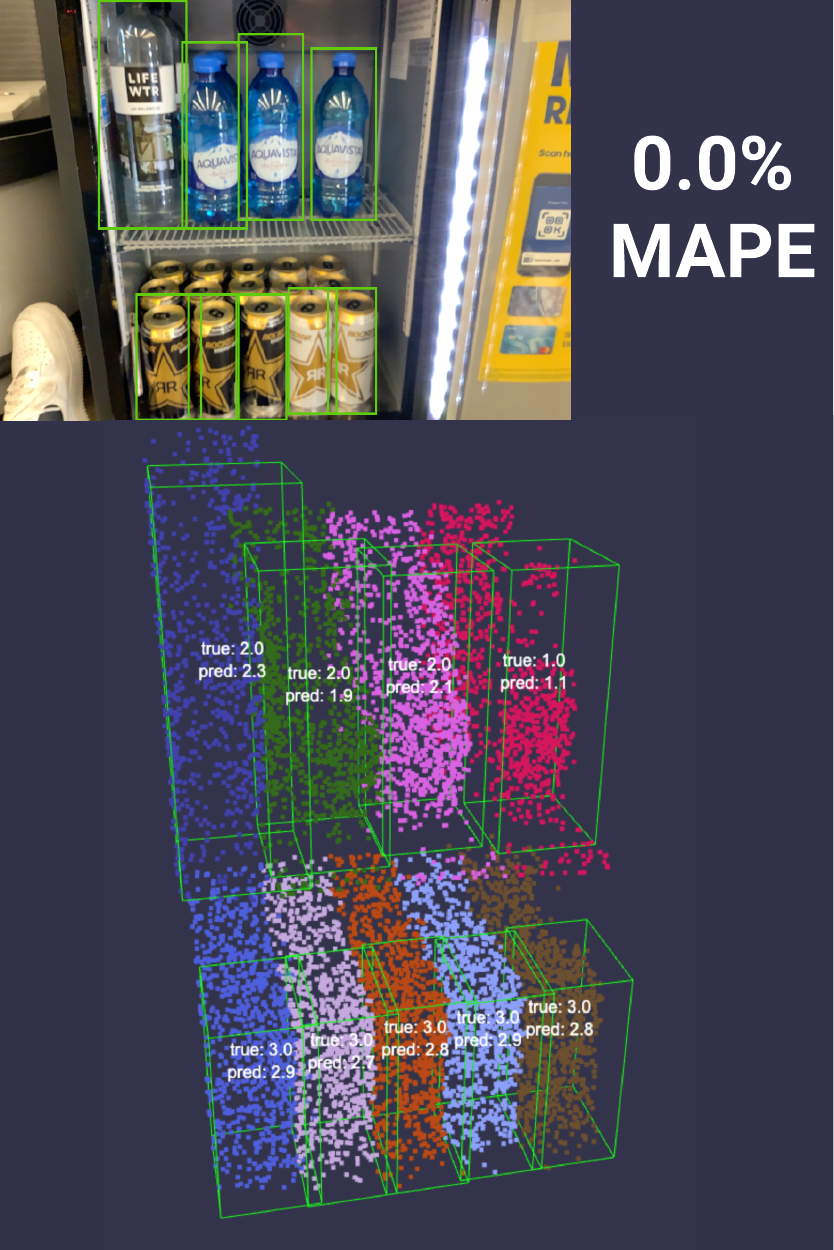} \\
    
     \end{tabular}
    \caption{\textbf{Example predictions from CountNet3D on real-world data}. (Best viewed in color with zoom in). For each scene we show the image, the PointBeam proposals, the ground truth, and the predicted counts. We also display the scene-level MAPE (after rounding). Each beam is given a  randomly generated color to highlight the beam regions.}
    \label{fig:pics}
\end{figure*}

CountNet3D is trained in two stages. First, the YOLOv5 model is trained on image data alone to detect finegrained beverage products. This model consists of 567 fine grained classes. Here, fine grained class refers to a very specific product classification such as \texttt{coca\_cola\_20oz\_bottle} or \texttt{coca\_cola\_1L\_bottle}. Detection results are reported in Table \ref{tab:yolo}. Second, the YOLOv5 weights are fixed and the network processing the point cloud is trained to regress object counts in a focused sub-volume of the scene. The predictions are aggregated across sub-volumes to the class-level to output a count estimate for each product. The object count regression results are reported in Table \ref{tab:countnet}. Example outputs of our CountNet3D implementation are presented in Figure \ref{fig:pics}. 

The visual layer is trained with a proprietary dataset comprised annotated of images and point clouds. Each point cloud is densely labelled; small regions are annotated with the product type and ground truth object counts. The dataset construction is very similar to the open source version in \cite{jenkins2023wacv}.

\begin{table}[]
    \centering
    \begin{tabular}{ccccc}
         &  mAP50-95 & mAP50 & Precision & Recall \\
         \toprule
         YoloV5 & 0.793 &  0.927 & 0.897 & 0.901
    \end{tabular}
    \caption{YoloV5 training results on 567 finegrained classes.}
    \label{tab:yolo}
\end{table}

\begin{table}[]
    \centering
    \begin{tabular}{cccc}
         &  MAE & MSE & MAPE \\
         \toprule
         CountNet3D & 1.280 &  5.672 & 0.077
    \end{tabular}
    \caption{Object count regression results from CountNet3D. We report Mean Absolute Error (MAE), Mean Squared Error (MAE) and Median Absolute Percentage Error (MAPE).}
    \label{tab:countnet}
\end{table}

\section{Additional Experiments and Results}

\begin{figure}
    \centering
     \includegraphics[width=0.4\textwidth]{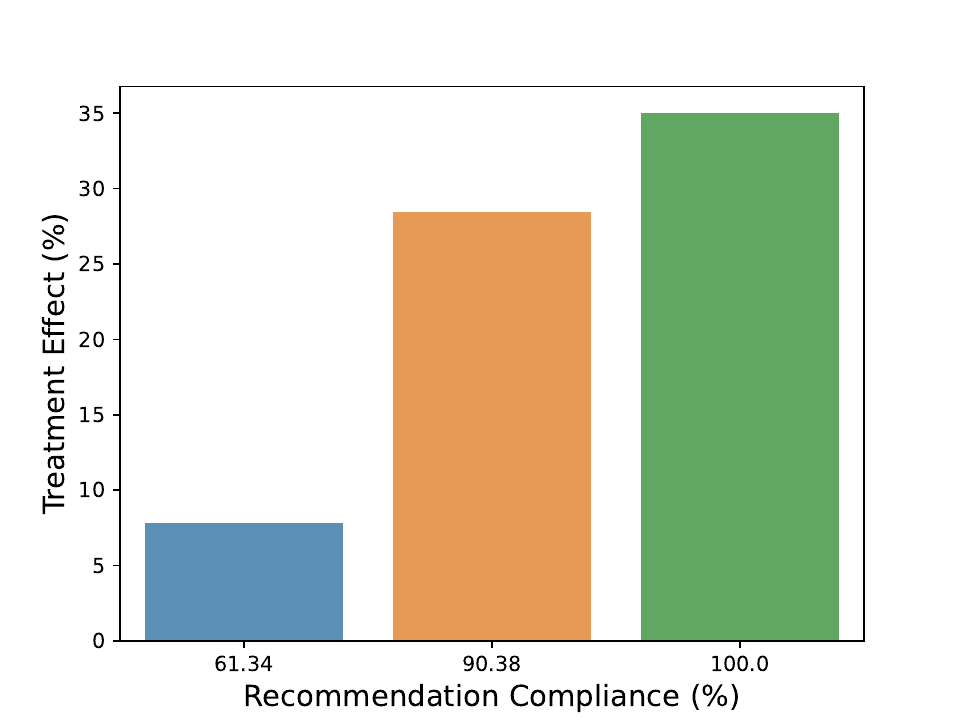}
     \caption{Mean treatment effect as a function of recommendation compliance.}
     \label{fig:compliance}
\end{figure}

\subsection{Details of Offline Experiments}

\subsubsection{Dataset Construction}
As discussed in the paper, we use the offline policy evaluator proposed by \cite{li2011unbiased} to benchmark different search algorithms against a static dataset. The dataset is collected from user logs taken from the control group in experiment \#1. The control group saw no recommendations during the A/B test is therefore unbiased w.r.t the search algorithms benchmarked. 

For all offline and online experiments, we use data collected by our system. The datasets in our offline experiments and online deployment studies are all comprised of longitudinal data describing daily sales for each product on display. The datasets all have the following schema:

\begin{itemize}
    \item store id: identifier of the store
    \item display id: identifier of the product display
    \item scanned datetime: timestamp of the user’s interaction with the display
    \item product id: identifier of observed product
    \item timedelta: difference (in hours) since previous interaction with the display
    \item sales: estimated numbers of units sales (from CountNet3D)
\end{itemize}

\subsubsection{Benchmarks}
For the $\epsilon$-greedy implementation, we set $\epsilon$ = 0.1. For the genetic algorithm we randomly marry 70\% of pairs together to form a new configuration, and choose a random action 30\% of the time.

Both Deep Ensembles and MOPO rely on ensembles of neural networks to learn a probabilistic model of the environment. In each case, we train 5 fully connected neural networks with three hidden layers with dimensions $[256, 256, 256]$ and 135 input features. The input features include a one hot vector for each product, the number of observed product facings, and day of the week features. Each network is randomly initialized and trained for 1000 epochs to minimize Gaussian negative log likelihood loss. In the case of the Deep Ensemble we use the same heuristic search algorithm desribed in Algorithm \ref{alg:search}. For MOPO, we perform 5 rollouts with the dynamics ensemble at each step, and update a tabular policy with a buffer of real observations and rollouts.

\subsection{Details of Online Experiments}

\subsubsection{Users}

\begin{table}[]
    \centering
    \begin{tabular}{ccc|c}
         \toprule
         &  \textbf{Control} & \textbf{Treatment} & \textbf{Treat. - Control}  \\
         \toprule
         Pre & 5.40 &  5.82 &  0.42\\ 
         Post & 4.40 & 6.77 &  2.37 \\ 
         \midrule
         Change  & -1.00 (-18.61\%)  & +0.95 (+16.42\%) & $\textbf{+1.95 (+35.03\%)}^{\ast\ast}$\\
    \end{tabular}
    \caption{Experiment 1 Difference-in-Difference Results. We present daily average products sold, and their percentage changes. We use $^{\ast \ast}$ to denote significance at $\alpha = 0.05$.}
    \label{tab:did-exp1}
\end{table}

The core user in our online experiments is referred to in the industry as a merchandiser. His or her primary function is to ensure that shelves are replenished with product. In some cases, they are asked to make decisions about which products to place on shelves to improve sales. However, this process is usually driven by intuition, and most merchandisers lack concrete data to inform their decisions. Typically, the merchandiser surveys and restocks each display in a store on a daily basis. 

Users interact with \ours through a mobile application while they are merchandising a store. As the user replenishes product each day, the merchandiser is instructed to view the mobile app to get suggested, or recommended, product restock decisions. The output of \ours is a set of products and their quantities to put on the shelf. One important aspect of this real-world recommendation problem is that the merchandiser has the option to execute or skip the recommendation. For example, a merchandiser may be rushed to fill a large number of shelves and skip the recommended action all together. This motivates our use of a compliance statistics $\rho.$

\begin{table}[]
    \centering
    \begin{tabular}{ccc|c}
         \toprule
         &  \textbf{Control} & \textbf{Treatment} & \textbf{Treat. - Control}  \\
         \toprule
         Pre & 12.74 & 9.50 &  -3.24 \\ 
         Post & 10.86 & 10.74 &  -0.12 \\ 
         \midrule
         Change  & -1.88 (-14.76\%)  & +1.24 (+13.02\%) & $\textbf{+3.12 (+27.78\%)}^{\ast\ast}$\\
    \end{tabular}
    \caption{Experiment 2 Difference-in-Difference Results (High Compliance Group). We present daily average products sold, and their percentage changes. We use $^{\ast \ast}$ to denote significance at $\alpha = 0.05$}
    \label{tab:did-exp2}
\end{table}

\subsubsection{User Workflow}

The user workflow is fairly straightforward and integrates into their existing merchandising responsibilities. Prior to restocking the shelf, the user performs the pre-scan to get an updated inventory measurement and capture the current state, $\mathcal{S}$. The user then views the app, and queries the latest recommendations. At this point the user chooses whether or not to implement the recommendations and restocks the display. Finally the user completes the task by performing a post-scan and setting a baseline inventory.

\subsection{RBP Toy Experiments}

\begin{figure*}[tb]
    \centering
    \begin{tabular}{cccc}
        \includegraphics[width=0.3\linewidth]{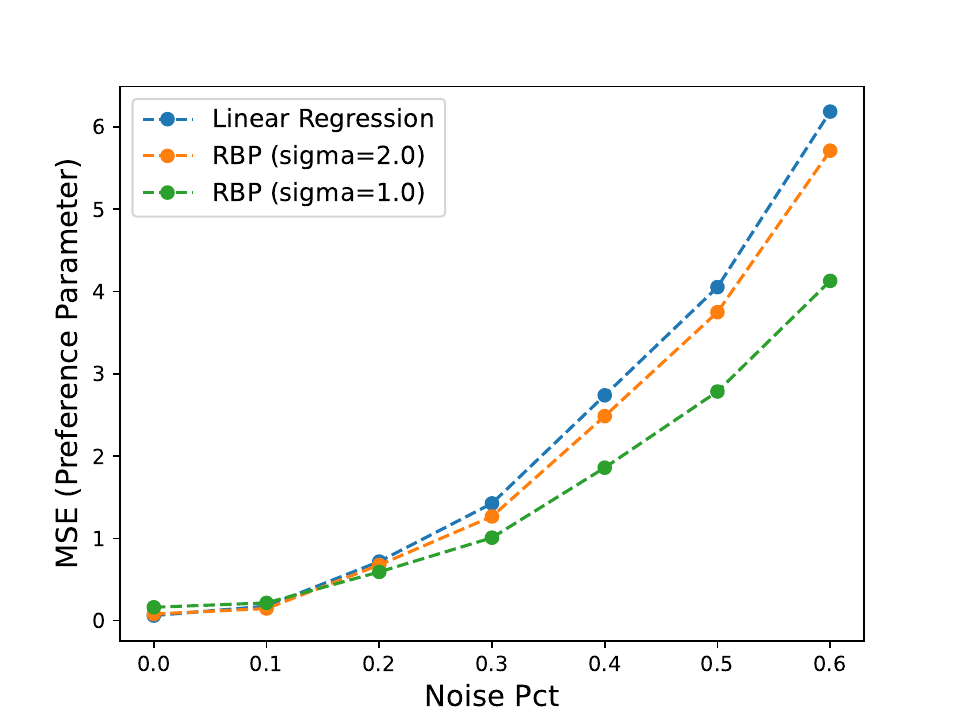}
        & \includegraphics[width=0.3\linewidth]{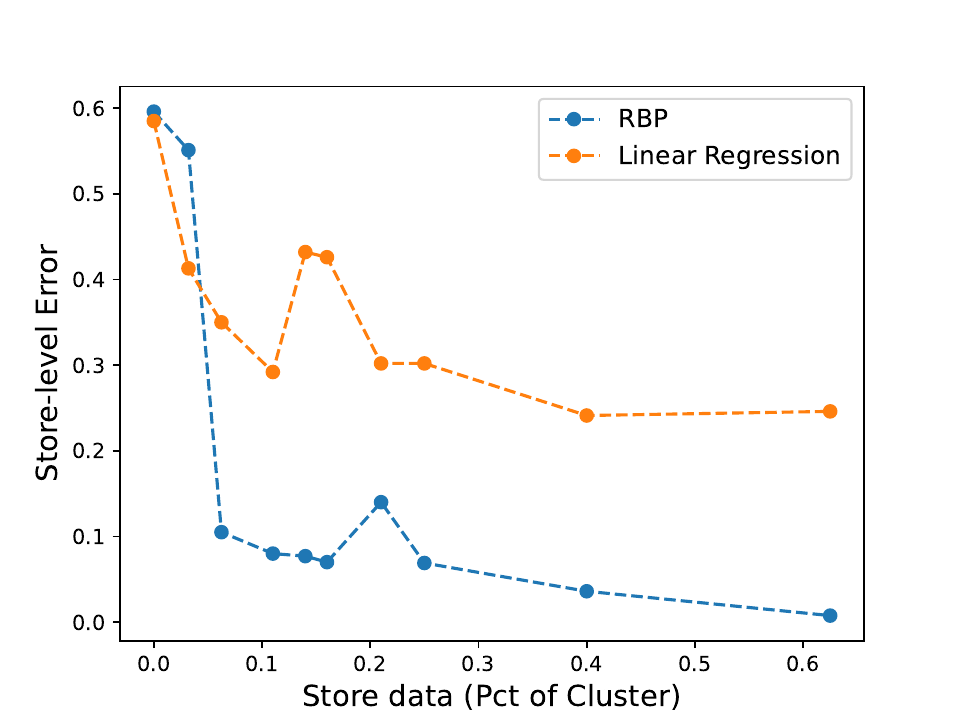}
        & \includegraphics[width=0.3\linewidth]{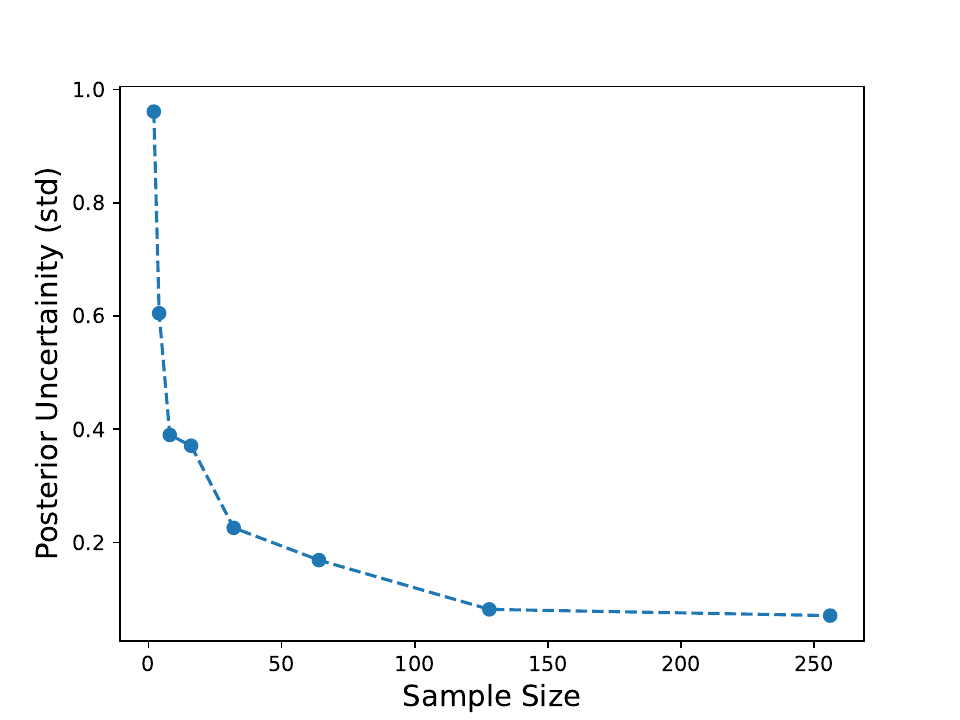}
    \end{tabular} \\

    \setlength{\tabcolsep}{20pt}
    \begin{tabular}{cccc}
        {\small(a) Robustness to Outliers}  & 
        {\small(b) Sample size of store-level error} & {\small (c) Sample size on Uncertainty}
        \end{tabular} \\

    \caption{Evaluation of the Robust Bayesian Payoff (RBP) model on a toy dataset. We demonstrate that RBP has three attractive properties over non-Bayesian approaches: (a) RBP is more robust to outliers. As the data noise is increased, RBP is more robust than linear regression; robustness increases as the prior strength increases (smaller $\sigma$). (b) As more data is acquired within a store, RBP discounts information from other stores in the cluster. Linear Regression equally weights all data, increasing store-level error. (c) In general, posterior uncertainty decreases as more data is acquired.  }
    \label{fig:toy-exp}

\end{figure*}

We claim that RBP has three advantages over non-Bayesian approaches. First, RBP is more robust to outliers and data noise. This is due to the robust, Laplacian priors on the store-level $\beta_{i, l, k}$ parameters. Laplacian priors are known to be more robust to noise and outliers than Gaussian priors. Second, RBP can effectively trade-off information from the cluster and the store through it's hierarchical structure. As more data is acquired within a store, RBP discounts information from other stores in the cluster. Conversely, linear regression and neural nets equally weight all data, increasing store-level error. Third, posterior uncertainty decreases as more data is acquired. We test these three claims in a toy environment on simulated data where we know the ground truth parameter values and can manipulate observation noise and dataset size. We report our results in Figure \ref{fig:toy-exp}.

We see all three claims validated. In Figure \ref{fig:toy-exp} (a), we see RBP is more robust as noise is increased. We can also control robustness through the hyperparameter $\sigma_0$ and $\sigma_1$ parameters. In Figure \ref{fig:toy-exp} (b) we see that RBP better learns store-level preferences, and the error decreases as a function of more in-store data. Conversely, linear regression flat lines and cannot weigh store-level data against cluster-level data. In Figure \ref{fig:toy-exp} (c), we see the posterior uncertainty decreases as more data is observed.  

\subsection{Selecting the Number of Clusters}
We choose the number of clusters using an auxiliary, 180-day snapshot of store-level sales for 897 different products and 2,548 stores. We start by choosing a proposed number of clusters $k$. We then run SpAGMM + K-means to generate $k$ clusters of stores and group the store-level sales data by cluster. For each product we then perform a simple Analysis of Variance (ANOVA) test to determine if the product preferences vary across clusters. If the difference across groups is statistically meaningful ($\alpha=0.05$) we record \texttt{true}, otherwise we record \texttt{false}. Finally, we compute the proportion of products that recorded \texttt{true} to assess the cluster fit quality. We find $K=20$ is a roughly the point where adding another cluster yields a small increase fit, and set $K=20$ in our experiments.

\begin{figure*}
     \centering
     \begin{subfigure}[b]{0.435\textwidth}
         \centering
         \includegraphics[width=\textwidth]{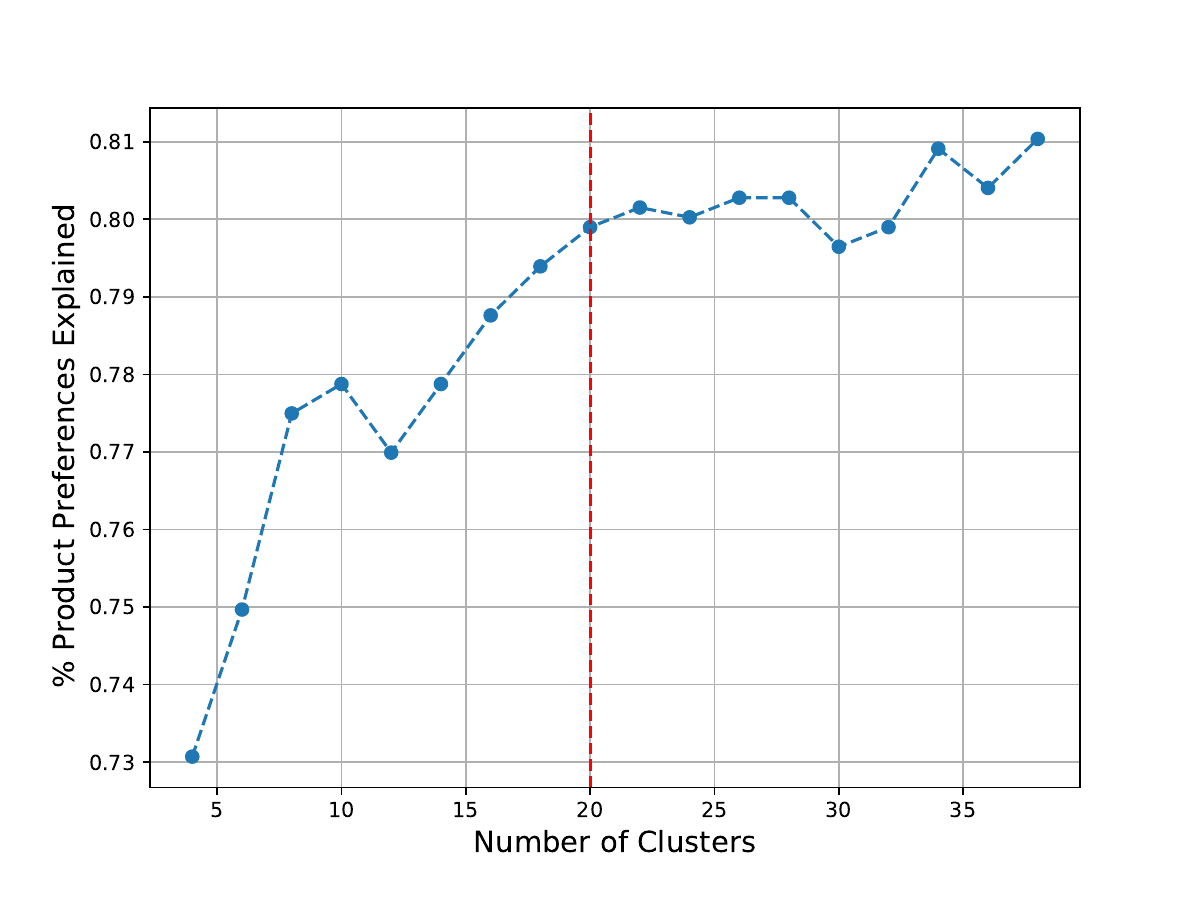}
         \caption{SpAGMM Cluster selection}
         \label{fig:cluster_a}
     \end{subfigure}
     \begin{subfigure}[b]{0.5\textwidth}
         \centering
         \includegraphics[width=\textwidth]{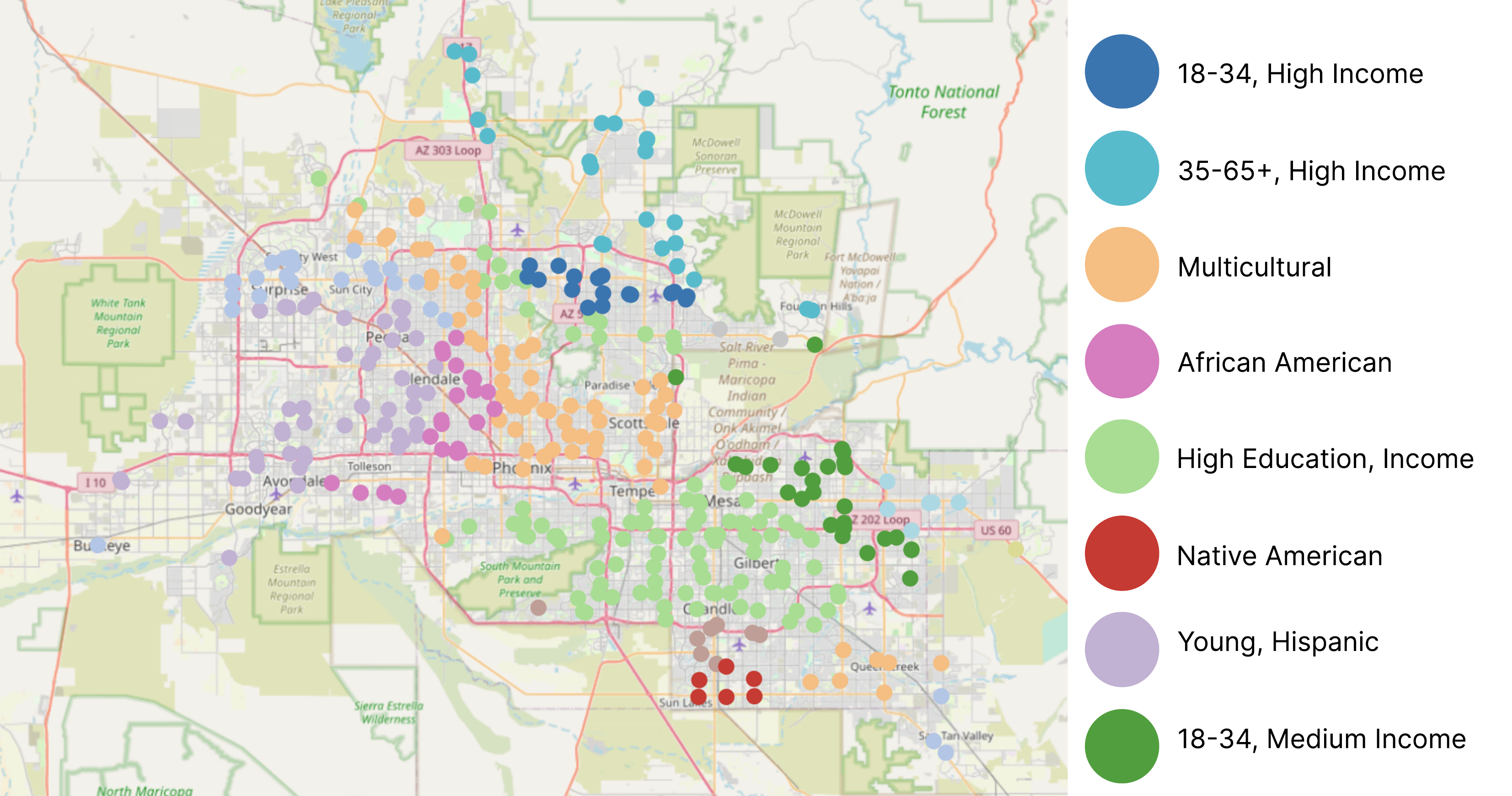}
         \caption{Learned store cluster assignments in Phoenix, Arizona, USA.}
         \label{fig:cluster_b}
     \end{subfigure}
     \caption{Diagnostics of our proposed clustering procedure. (a) Using an auxiliary sales dataset, we plot the percentage of product preferences explained by the learned clusters, against the number of clusters. We perform a series of product-level ANOVA tests to calculate the y-axis. We choose $K=20$ as it balances data fit and interpretability. (b) We show learned store-level cluster assignments in Phoenix, AZ. We discover interesting local neighborhoods, typical of urban data.}
     \label{fig:cluster}
\end{figure*}
    
\subsubsection{Visualization of Clusters}

In Figure \ref{fig:cluster_b} we show a 2D plot of the learned store-level clusters, after applying a TSNE dimensionality reduction. We note that the inter-cluster variance is generally quite low. This indicates that stores within a cluster tend to be demographically similar. Some clusters (i.e, \#13) are comprised of nearly identical stores, while others exhibit more variation (i.e., \#9). We also note high intra-cluster variance. Meaning, clusters are cleanly separated and don't overlap. These properties are indicate that the clusters fit the data, and meaningful demographic patterns have been discovered.

Additionally, Figure \ref{fig:cluster_b} depicts the store-level cluster assignments of stores in Phoenix, Arizona, along with a high-level summary of various clusters identified by SpAGMM + K-means. First, we observe clear spatial stratification, which is typical of urban data.  We note highly educated people living near Mesa and Tempe, home of Arizona State University. Additionally, see a more diverse racial group near downtown, and a higher Hispanic concentration towards Glendale. Our primary conclusion from Figure \ref{fig:cluster_b} is that we can detect spatial variation in demographic profiles, and that they do explain variation in product preferences.

\subsection{Product Case Study}

\begin{figure*}
     \centering
     \begin{subfigure}[b]{0.55\textwidth}
         \centering
         \includegraphics[width=\textwidth]{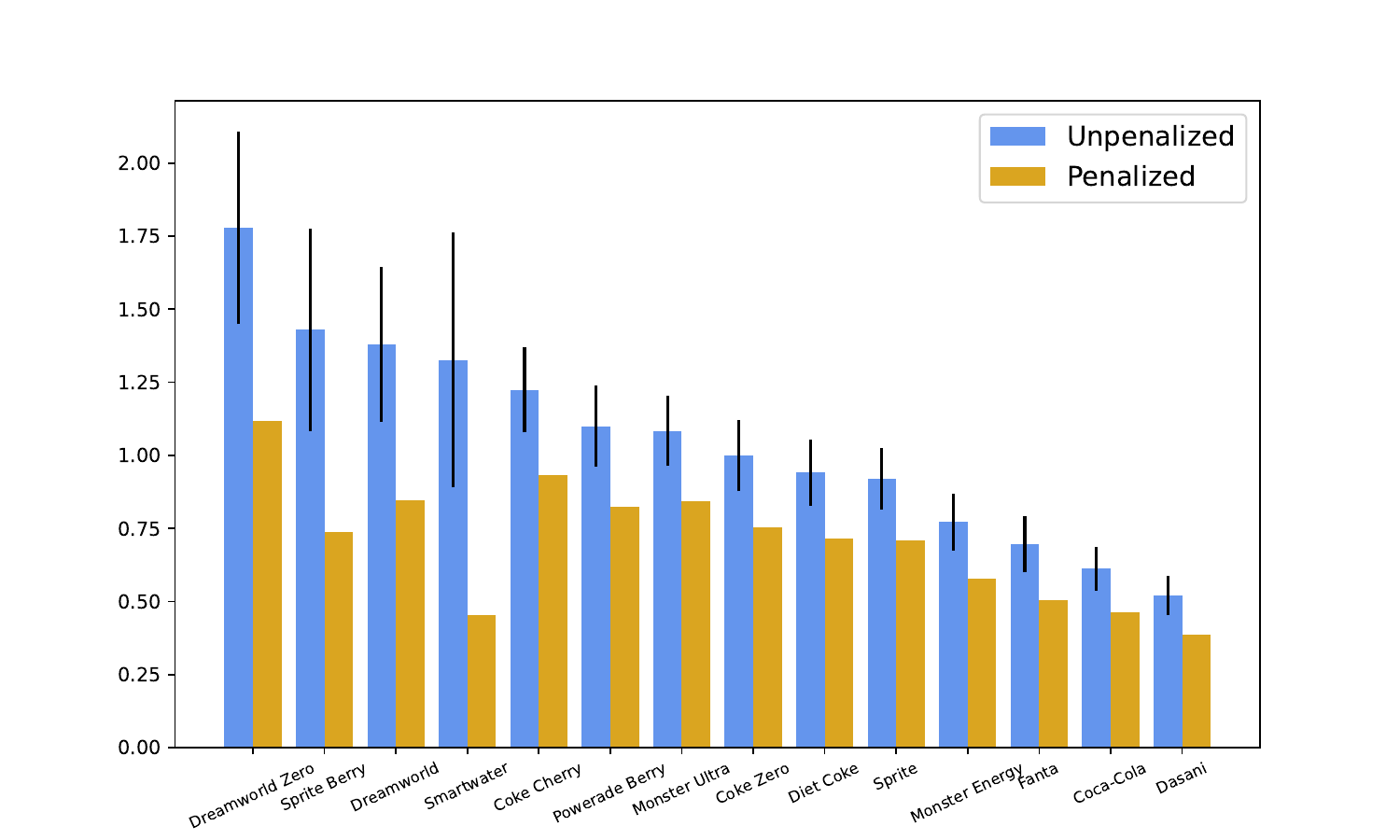}
         \caption{Product Case Study}
         \label{fig:epf}
     \end{subfigure}
     \hfill
     \begin{subfigure}[b]{0.35\textwidth}
         \centering
         \includegraphics[width=\textwidth]{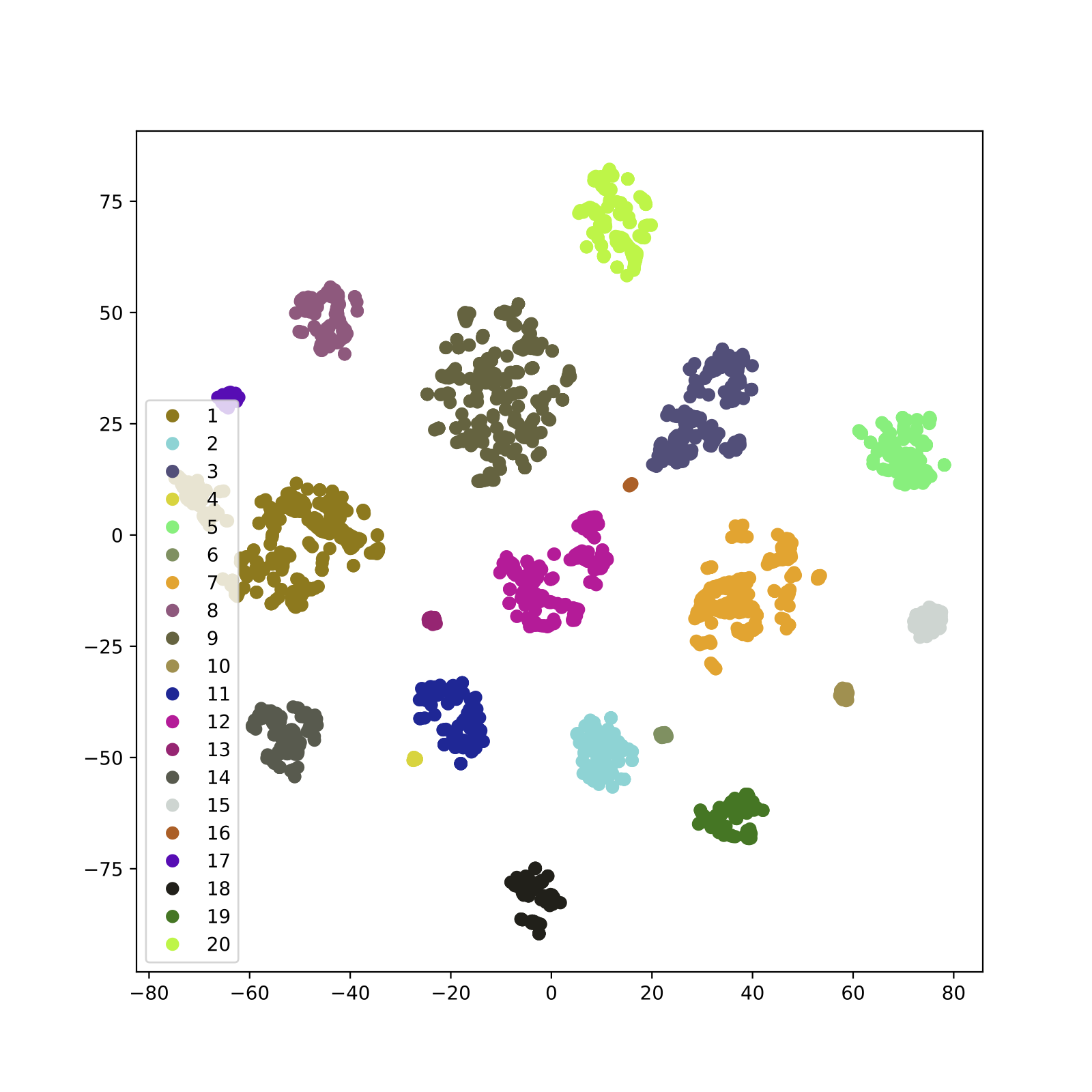}
         \caption{TSNE Visualization of Spatial Cluster}
         \label{fig:tsnse}
     \end{subfigure}
     \caption{ (a) Penalized and unpenalized PEPF statistics from an example display taken from experiment \#2. (b) A two dimensional TSNE visualization of the 20 store clusters learned in our experiments. We see clearly delineated, and relatively tight clusters in the 2D space, indicative of good data fit. }

\end{figure*}

In Figure \ref{fig:epf} we present an example product display observed in field experiment \#2. The top 14 candidates for this display are presented along with the unpenalized (blue) and penalized (gold) reward estimates, or PEPF scores. Products are sorted by unpenalized reward. We also plot the posterior standard deviation of reward with black vertical lines. We set the penalty coefficient, $\lambda=2.0$.

From this penalized and unpenalized comparison, we make the following conclusions. First, Many of the products with high expected reward also have high uncertainty. This is consistent with our rationale for proposing RBP. We want to avoid overly optimistic actions, and high reward estimates could in fact be spurious. Second, penalizing by uncertainty changes the product order. For example, \texttt{Sprite Berry} and \texttt{Smartwater} are two of the top four product by unpenalized PEPF. When we apply the uncertainty penalty, their rank shifts downward. In the case of \texttt{Smartwater}, it's uncertainity is so high, that it becomes the lowest ranked product of the candidate set. In this example, \ours removes space from the bottom two products \texttt{Coca-cola} and \texttt{Dasani} and allocates it to \texttt{Dreamworld} and \texttt{Dreamworld Zero}, because their penalized rewards estimates are higher.


\begin{figure*}
     \centering
     \begin{subfigure}[b]{0.45\textwidth}
         \centering
         \includegraphics[width=\textwidth]{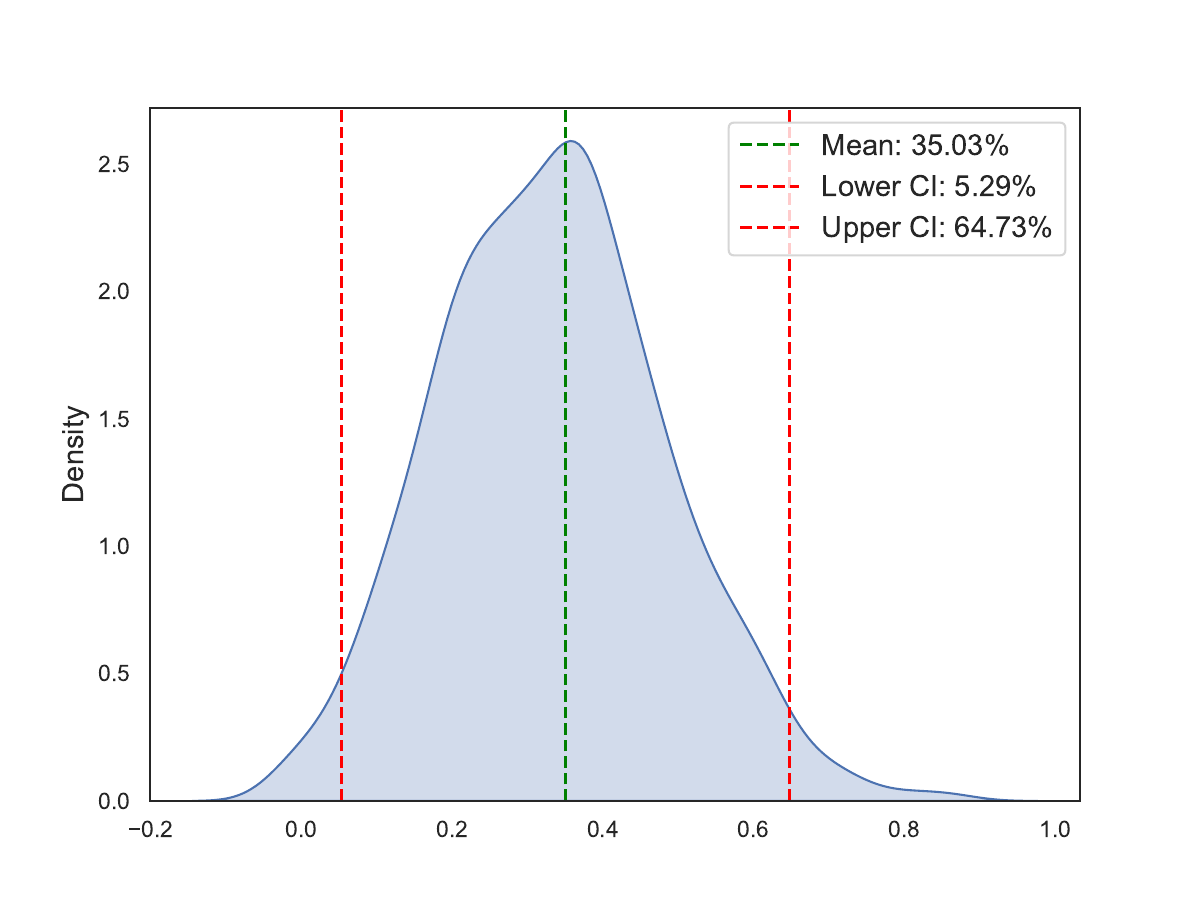}
         \caption{Exp 1: DID effect (all Stores)}
         \label{fig:res_a}
     \end{subfigure}
     \hfill
     \begin{subfigure}[b]{0.45\textwidth}
         \centering
         \includegraphics[width=\textwidth]{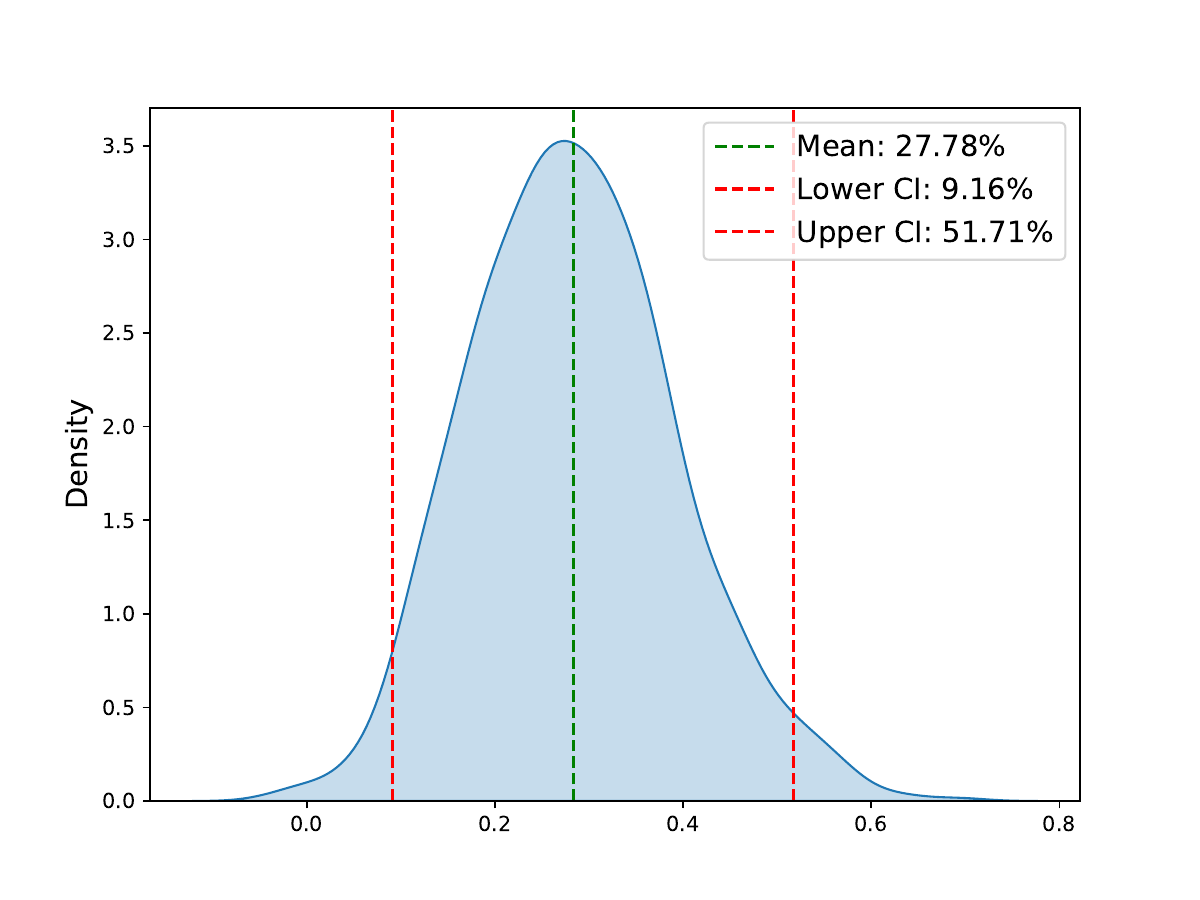}
         \caption{Exp 2: DID effect (high compliance)}
         \label{fig:res_b}
     \end{subfigure}
     \caption{Summary of our results. (a) The DID treatment effect for experiment 1, with 95\% CI; (b) the treatment effect of the high compliance group in experiment 2, with 95\% CI (b); (c) the mean treatment effect as a function of recommendation compliance. }
     \vspace{-5mm}
\end{figure*}

\end{document}